\documentclass{article}

\PassOptionsToPackage{numbers, compress}{natbib}

\usepackage[main, final]{neurips_2026}

\usepackage[utf8]{inputenc} 
\usepackage[T1]{fontenc}    
\usepackage{hyperref}       
\usepackage{url}            
\usepackage{booktabs}       
\usepackage{amsfonts}       
\usepackage{nicefrac}       
\usepackage{microtype}      
\usepackage{xcolor}         
\usepackage{amsmath}
\usepackage{amssymb}
\usepackage{graphicx}
\usepackage{subcaption}
\usepackage{wrapfig}
\usepackage{booktabs} 
\usepackage{multirow} 
\usepackage{enumitem}
\usepackage{bm}
\usepackage{titlesec}
\usepackage{etoc}
\usepackage{makecell}

\titlespacing*{\section}{0pt}{1.2ex plus .2ex minus .2ex}{0.8ex plus .2ex}
\titlespacing*{\subsection}{0pt}{1.0ex plus .2ex minus .2ex}{0.6ex plus .2ex}
\titlespacing*{\paragraph}{0pt}{0.5ex plus .1ex minus .1ex}{1em}
\title{\textsc{DiagNO}: Diagonal Spherical Neural Operators for Heterogeneous Earth Dynamics Modeling}

\author{%
{\normalfont
\textbf{Herui Li$^{1}$, Bin Lu$^{1,*}$, Haonan Qi$^{1}$, Lei Zhou$^{2}$,
Luoyi Fu$^{3}$, Xinbing Wang$^{1}$, Meng Jin$^{4,*}$}}\\[1.0ex]
{\normalfont
$^{1}$School of Information Science and Electronic Engineering, Shanghai Jiao Tong University}\\
{\normalfont
$^{2}$School of Oceanography, Shanghai Jiao Tong University}\\
{\normalfont
$^{3}$School of Computer Science, Shanghai Jiao Tong University}\\
{\normalfont
$^{4}$School of Artificial Intelligence, Shanghai Jiao Tong University}
}

\begin{document}

\maketitle
\begingroup
\renewcommand{\thefootnote}{\fnsymbol{footnote}}
\footnotetext[1]{Corresponding authors: Bin Lu (\href{mailto:robinlu1209@sjtu.edu.cn}{robinlu1209@sjtu.edu.cn}) and Meng Jin (\href{mailto:jinm@sjtu.edu.cn}{jinm@sjtu.edu.cn}).}
\endgroup

\begin{abstract}
Accurate forecasting of Earth's dynamic systems is vital in geophysics, where neural operators offer a powerful data-driven paradigm. Standard operators often treat the Earth as a flat Euclidean plane, introducing geometric artifacts. In contrast, Spherical Neural Operators (SNOs) resolve this mismatch by operating within the spherical spectral space, naturally adapting to spherical geometries. However, for computational efficiency, existing SNOs typically rely on the rotation equivariance assumption. This assumption simplifies the transitions of physical states into isolated spectral modal evolutions, thereby over-idealizing the Earth as a homogeneous system. 
To solve this problem, we no longer rely on the rotation equivariance assumption that isolates spectral modal evolution, but directly derive an explicit modal interaction kernel and propose \underline{Diag}onal Spherical \underline{N}eural \underline{O}perators named \textsc{DiagNO}. 
\textsc{DiagNO} explicitly parameterizes cross-modal interactions through spectral modal decomposition, thereby preserving the inherent heterogeneity of Earth dynamics. Furthermore, leveraging the unique \textit{diagonal} structure of the spherical spectrum, we design inter- and intra-diagonal interaction mechanisms to natively capture essential zonal and meridional dynamics, respecting intrinsic geophysical regularities. Extensive experiments across simulated fluid dynamics and real-world Earth system reanalysis (including atmosphere and ocean) datasets demonstrate \textsc{DiagNO}’s superior capability, consistently achieving state-of-the-art performance in multi-step forecasting. Beyond this, we validate \textsc{DiagNO}’s practical analytical value in professional geophysical domains by assessing its fidelity in capturing climatological anomalies and maintaining physical consistency.
\end{abstract}

\section{Introduction}
Accurate forecasting of Earth's dynamic systems is vital in geophysics, particularly within oceanic and atmospheric sciences~\cite{Rose_Yu_1, fuxi_25, XiaomengHuang1, essd-18-5069-2026, KDD_BinLu}. While numerically solving governing partial differential equations (PDEs) is computationally prohibitive~\cite{numerical_prediction}, neural operators~\cite{FNO_21,DeepONET_21} offer a powerful data-driven paradigm, modeling the spatiotemporal evolution of physical fields via functional space mappings.

However, standard neural operators encounter a topological hurdle when applied to global-scale Earth systems: they treat the spherical manifold as a Euclidean plane. Such planar modeling does not account for the intrinsic curvature and coordinate singularities of sphere~\cite{staniforth2012horizontal}, inevitably introducing geometric mismatches. Consequently, it triggers radial artifacts near the poles that can completely overwhelm the valid prediction. To resolve the mismatch, spherical neural operators, such as SFNO~\cite{SFNO_23} and its variants~\cite{GSNO_26,SMgNO}, shift the framework to the spherical spectral domain, naturally adapting to spherical geometries (we present a detailed analysis of this phenomenon in Section \ref{artifacts}).

Despite resolving the mismatch through spherical spectral modeling, existing spherical neural operators rely on the assumption of rotation equivariance~\cite{Spherical_cnns} for computational efficiency, which mathematically simplifies state transitions into isolated spectral modal evolutions (detailed in Appendix \ref{proof:isolated_modes}). By enforcing uniform rotational relationships, this approach \emph{over-idealizes the Earth as a homogeneous system}. However, the Earth is a \emph{heterogeneous dynamical system}, where physical evolution is tied to absolute geographical locations rather than relative rotational symmetries~\cite{torsvik2014deep, dziewonski2010mantle}. This heterogeneity is driven by non-uniform boundary conditions, such as continental geometries~\cite{complex_cg}, and spatially varying parameters like Coriolis force~\cite{Coriolos_force}. Consequently, the intrinsic heterogeneity of the Earth system cannot be adequately characterized by over-idealized rotation equivariance.

\begin{wrapfigure}{r}{0.40\textwidth} 
  \vspace{-10pt}
  \centering  \includegraphics[width=0.38\textwidth]{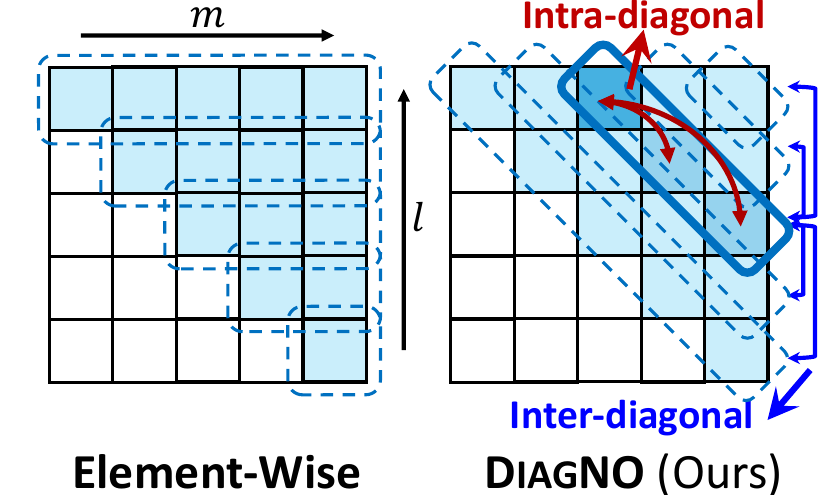}
  \caption{\textbf{Comparison in spherical spectral space.} Spherical Harmonics form a triangular $(l, m)$ spectrum. Unlike previous SNOs using element-wise transforms, \textsc{DiagNO} enables intra- and inter-diagonal interactions.}
  \label{fig:comparison}
  \vspace{-15pt}
\end{wrapfigure}
To address the limitation, we propose the Diagonal Spherical Neural Operator named \textbf{\textsc{DiagNO}}\footnote{The \textsc{DiagNO} resource is available at \href{https://github.com/jingwei-sjtu/DIAGNO}{https://github.com/jingwei-sjtu/DIAGNO}.}, which shifts the modeling paradigm from isolated modal evolution to explicit cross-modal interaction. Rather than relying on idealized system modeling, \textsc{DiagNO} utilizes direct spectral modal decomposition to derive a parameterized interaction kernel, thereby effectively capturing the heterogeneity of the Earth system. Furthermore, we observe that Spherical Harmonic spectrum inherently characterizes the distribution of zonal and meridional dynamics. Therefore, instead of restricting spectral modes to isolated element-wise evolutions, \textsc{DiagNO} leverages this unique \textit{diagonal} distribution to enable cross-modal interactions within the spectral domain, which is illustrated in Figure \ref{fig:comparison}. This approach strictly respects the geophysical regularities of the Earth.

Extensive experiments across simulated fluid dynamics and various Earth system reanalysis datasets (including both atmosphere and ocean) demonstrate that \textsc{DiagNO} consistently achieves state-of-the-art performance in forecasting. Furthermore, we validate \textsc{DiagNO}’s practical analytical value within professional geophysical domains by assessing its fidelity in capturing climatological anomalies and its ability to maintain physical consistency. These results confirm that \textsc{DiagNO} provides a reliable paradigm for modeling the complex and heterogeneous dynamics of the Earth system. 
Overall, our contributions are summarized as follows:

\noindent $\bullet$ \quad Instead of relying on the idealized rotation equivariance assumption of previous spherical neural operators, which restricts physical state transitions to isolated modal evolutions, we model explicit cross-modal interactions through direct spectral modal decomposition, fundamentally preserving the inherent heterogeneity of Earth's dynamic systems.

\noindent $\bullet$  \quad We observe that the diagonal structure of the Spherical Harmonic spectrum naturally characterizes global-scale zonal and meridional dynamics. Leveraging this unique spectral property, we propose a diagonal-based interaction strategy that enables explicit intra- and inter-diagonal interactions, strictly respecting intrinsic geophysical regularities.

\noindent $\bullet$ \quad We comprehensively evaluate \textsc{DiagNO} across critical Earth dynamics scenarios, including atmospheric and oceanic forecasting, where it consistently achieves state-of-the-art (SOTA) results. Furthermore, we demonstrate the model's practical analytical value in professional geophysics, evidenced by its high fidelity in capturing climatological anomalies and its robust ability to maintain physical consistency.
\section{Related Works}
Neural operators provide a data-driven paradigm for solving PDEs in complex physical scenarios such as fluid dynamics. Unlike traditional numerical solvers that rely on computationally expensive discretization and iterative processes~\cite{numerical_prediction}, neural operators directly learn mappings between infinite-dimensional function spaces in a data-driven manner~\cite{FNO_21}. This framework is embodied in diverse architectures. Spectral methods~\cite{KDD_HaonanQi, FNO_21, A-FNO_21, U-FNO_22,F-FNO_23, U-NO_23,MG-TFNO_24} capture global dependencies via frequency-domain parameterization. Transformer-based architectures~\cite{Galerkin_21, HT-NET_22, OFormer_23, Factformer_23,GNOT_23, Transolver_24, 0N0_24, HaixuWu1} utilize attention mechanisms to dynamically weight localized spatial interactions. State-space models~\cite{S4_22, MambaNO_24,SS-NO_26, PGMNO_26, AFMO_26} integrate state-space structures to track the temporal evolution of fluid states. Meanwhile, graph neural operators~\cite{MGNO_20,GNO_20,AMG_25} construct graph frameworks to resolve multi-scale features. However, when applied to Earth system modeling, these architectures fail to respect the inherent spherical geometry. Operating within Euclidean planes, they introduce spatial aliasing that manifests as radial artifacts near the poles. 

In contrast, \textbf{Spherical Neural Operators (SNOs)}, such as SFNO~\cite{SFNO_23} and GSNO~\cite{GSNO_26}, shift the framework from Euclidean planes to spherical spectral domain. However, to prioritize computational efficiency, conventional SNOs primarily rely on rotation-equivariant spectral operations. By modeling the Earth as an idealized system, these approaches restrict spectral modes to isolated evolution, thereby failing to capture the heterogeneity of global dynamics. Instead of relying on such an idealized assumption, we decompose physical states into spectral modes to derive an explicit interaction kernel. Building upon this, we leverage the distinct diagonal structure of the spectrum to design a novel \textsc{DiagNO} module that effectively captures heterogeneous interactions across orthogonal directions.
\begin{figure} 
  \centering

  \begin{subfigure}{0.16\linewidth}
    \centering
    \includegraphics[width=\linewidth]{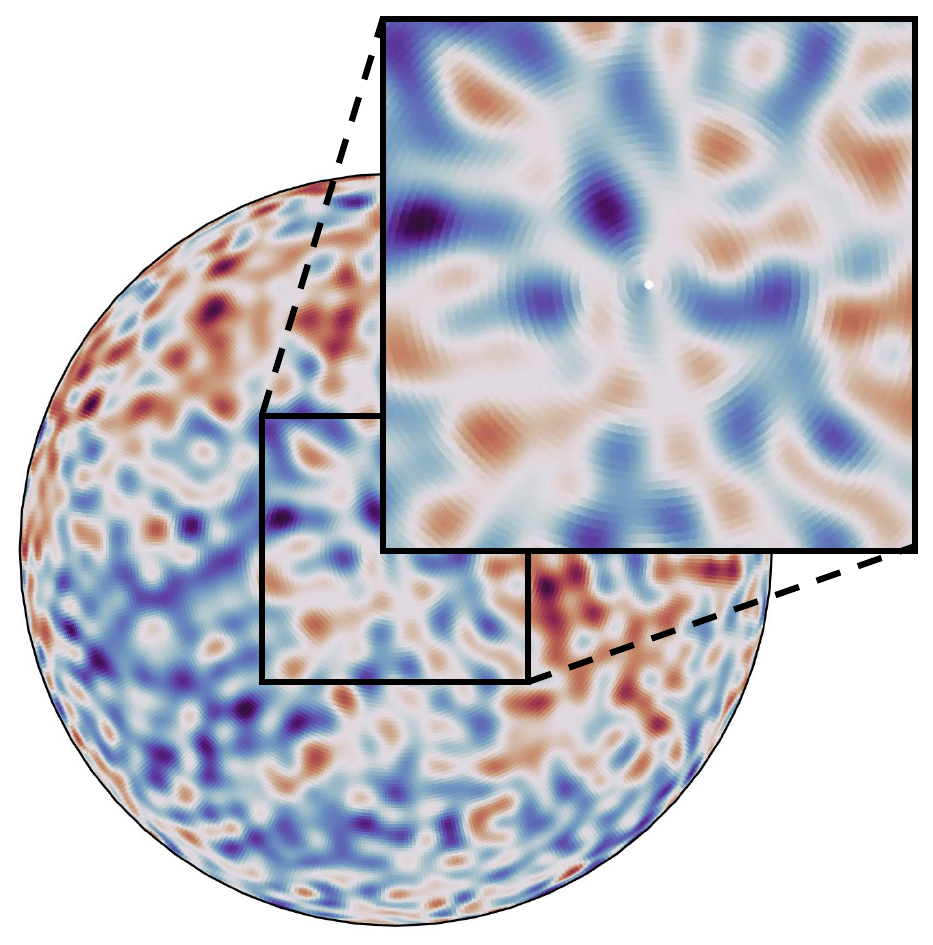}
    \caption{ground truth}
    \label{fig:fig2a}
  \end{subfigure}\hfill
  \begin{subfigure}{0.16\linewidth}
    \centering
    \includegraphics[width=\linewidth]{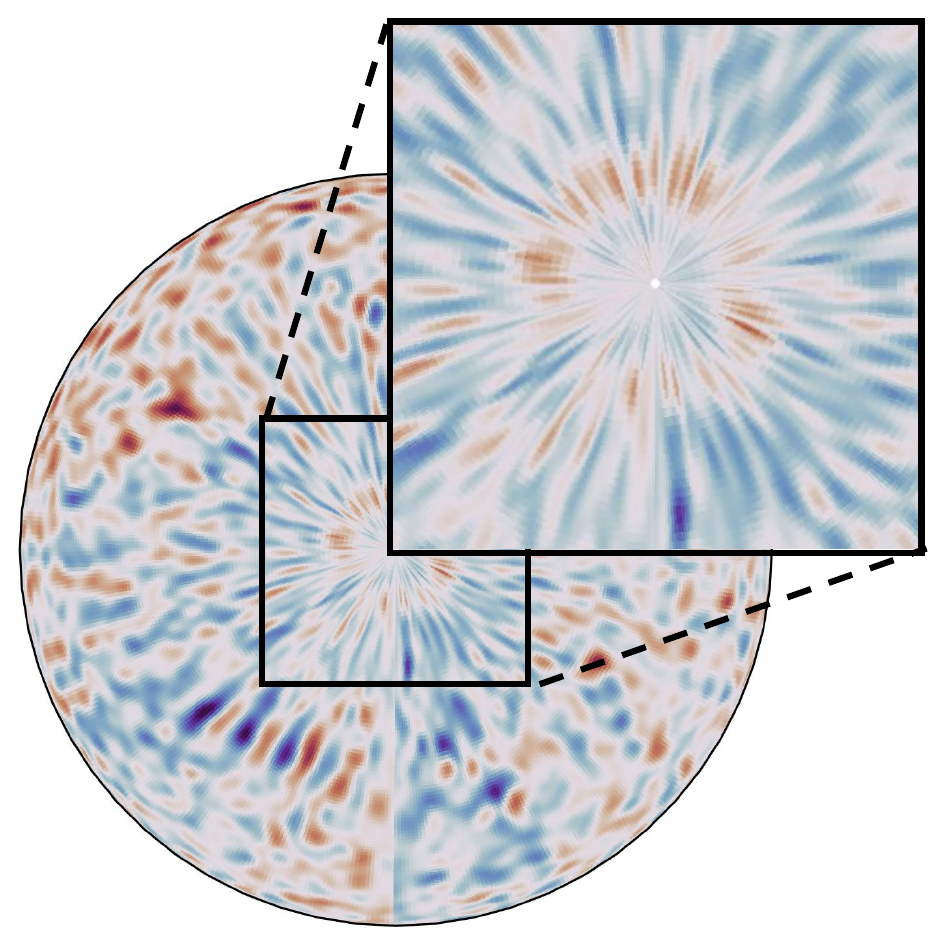}
    \caption{MambaNO}
    \label{fig:fig2b}
  \end{subfigure}\hfill
  \begin{subfigure}{0.16\linewidth}
    \centering
    \includegraphics[width=\linewidth]{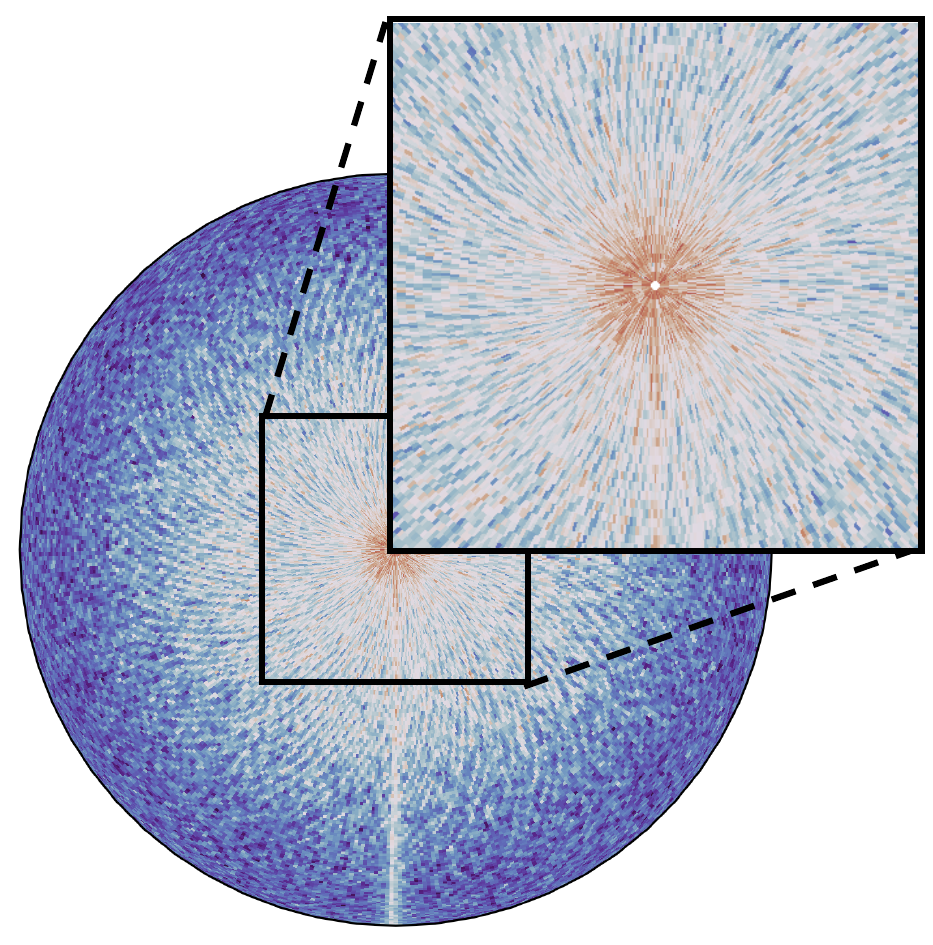}
    \caption{Transolver}
    \label{fig:fig2c}
  \end{subfigure}\hfill
  \begin{subfigure}{0.16\linewidth}
    \centering
    \includegraphics[width=\linewidth]{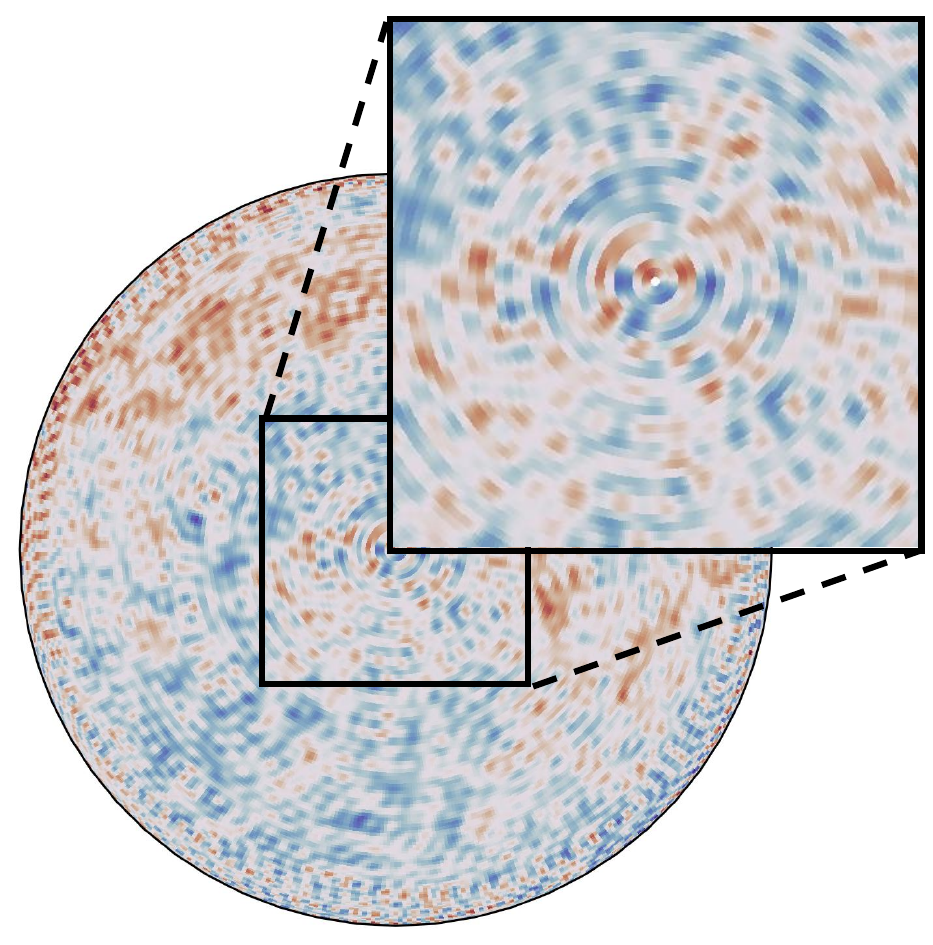}
    \caption{SFNO}
    \label{fig:fig2d}
  \end{subfigure}\hfill
  \begin{subfigure}{0.16\linewidth}
    \centering
    \includegraphics[width=\linewidth]{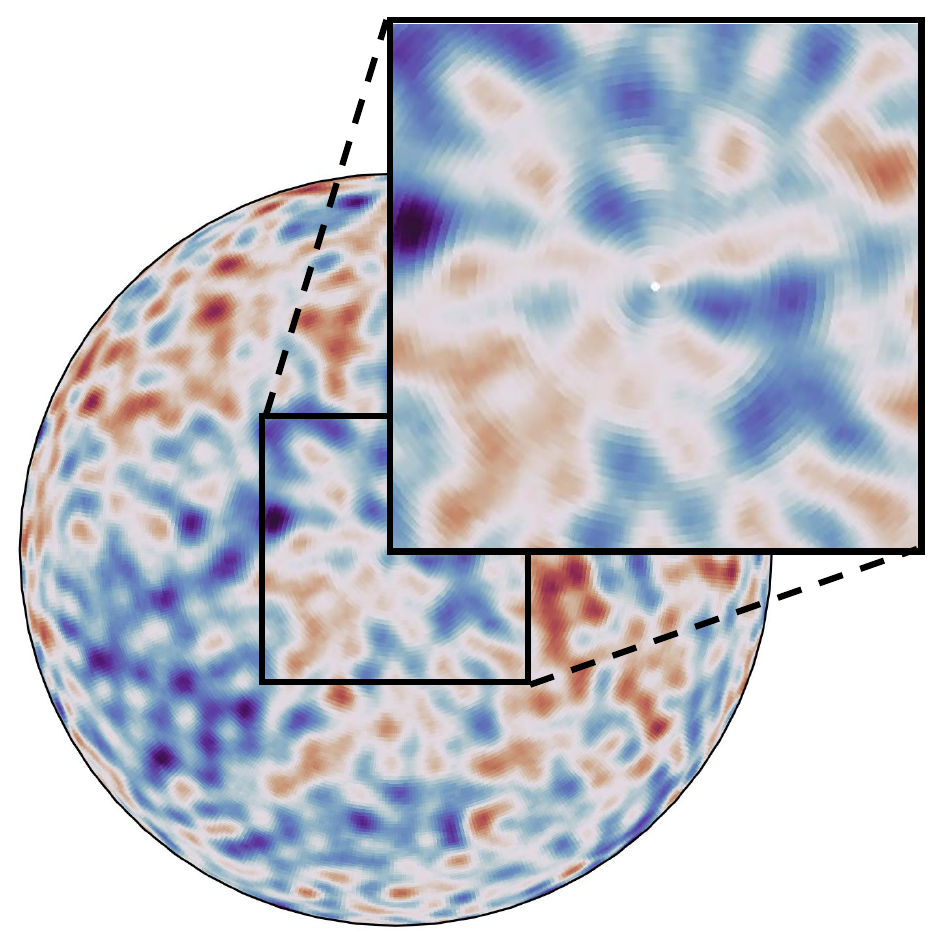}
    \caption{GSNO}
    \label{fig:fig2e}
  \end{subfigure}\hfill
  \begin{subfigure}{0.16\linewidth}
    \centering
    \includegraphics[width=\linewidth]{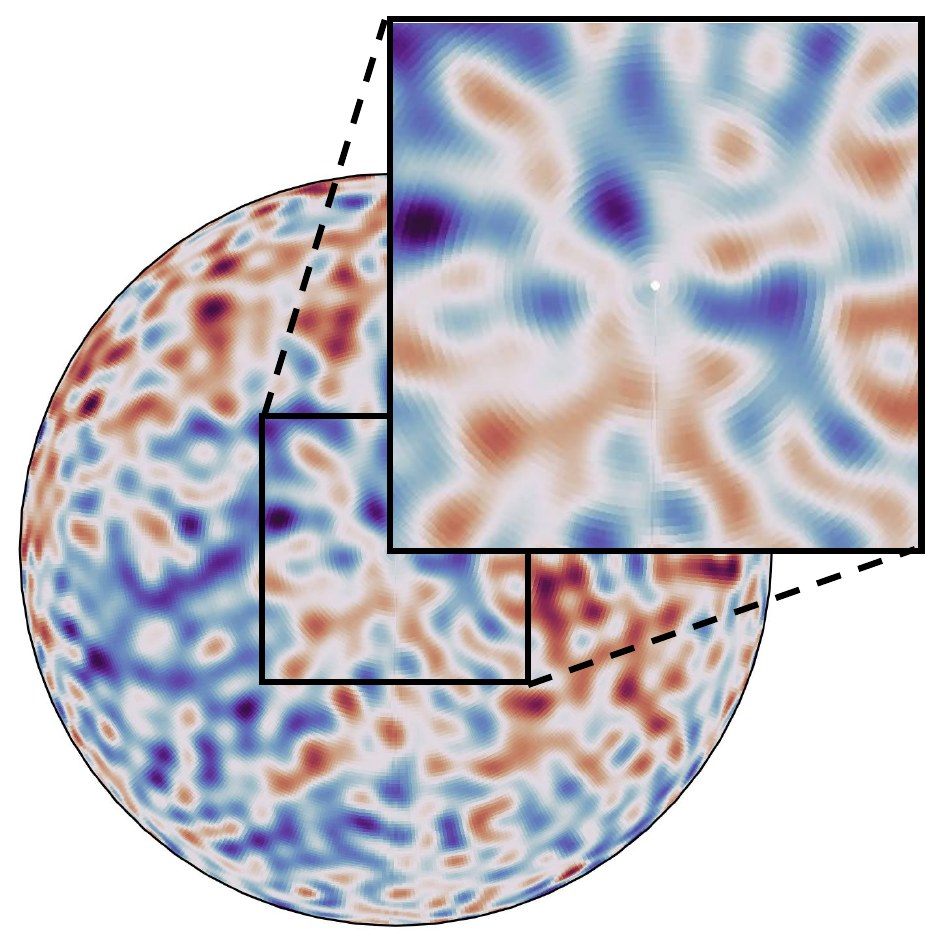}
    \caption{\textsc{DiagNO}}
    \label{fig:fig2f}
  \end{subfigure}

  \caption{\textbf{10-hour SSWE velocity field ($\bm{U}$) forecasting from a North Pole perspective.} \textbf{(a)} Ground truth. Standard modules \textbf{(b-c)} exhibit radial artifacts at the pole. SNOs \textbf{(d-f)} eliminate artifacts, though \textbf{(d-e)} suffer from energy dissipation. \textbf{(f)} \textsc{DiagNO} maintains the highest physical fidelity.}
  \label{fig:comparison_six}
  \vspace{-12pt}
\end{figure}
\section{Preliminaries}
\subsection{Problem Formulation}
\paragraph{Kernel Integral Operator.}
We consider the spatiotemporal evolution~\cite{YanLiu_1, YanLiu2} of physical dynamics governed by PDEs on a general spatial domain $\mathcal{D}$. Let $\mathbf{u}(x, t) \in \mathbb{R}^C$ denote the physical state at spatial coordinate $x \in \mathcal{D}$ and time $t$, where $C$ represents the number of characteristic channels. We seek to learn an operator $\mathcal{G}_\theta$ that approximates the true solution operator for single-step evolution:
\begin{equation}
    \mathbf{u}(x, t+\Delta t) = \mathcal{G}_\theta(\mathbf{u}(x, t)),
\end{equation}
where $\theta$ represents the learnable parameters and $\Delta t$ denotes the time interval. Following the neural operator paradigm, this mapping is formulated as a kernel integral operator~\cite{FNO_21}. Defining the input state as $\mathbf{u}(x) := \mathbf{u}(x, t)$ and the target state as $\mathbf{v}(x) := \mathbf{u}(x, t+\Delta t)$, the operator $\mathcal{K}$ is expressed as:
\begin{equation}
    \mathbf{v}(x) = (\mathcal{K}\mathbf{u})(x) = \int_{\mathcal{D}} \boldsymbol\kappa(x, x')\mathbf{u}(x')dx',
    \label{eq:integral}
\end{equation}
where $\boldsymbol\kappa(x, x') \in \mathbb{R}^{C \times C}$ is a continuous kernel representing interactions between $x$ and $x'$. Specifically, when modeling global geophysical dynamics, the domain $\mathcal{D}$ restricts to the two-dimensional sphere $\mathbb{S}^2$, with the coordinate specified as $x = (\theta, \phi) \in \mathbb{S}^2$, where $\theta \in [0, \pi]$ and $\phi \in [0, 2\pi)$.

\paragraph{Spherical Harmonic.}
For modeling on the sphere $\mathbb{S}^2$, Spherical Harmonics $Y_l^m(\theta, \phi)$ are typically adopted as the canonical orthogonal basis~\cite{Spherical_Harmonic}. The integer degree $l \geq 0$ and order $m \in [-l, l]$ collectively determine the spatial frequency of the basis function. Thus, the continuous spherical spatial field $\mathbf{u}(\theta, \phi)$ can be projected to the spectral domain through the Spherical Harmonic Transform (SHT)~\cite{SHT}. Specifically, the complex spectral coefficients $\hat{\mathbf{u}}_{l,m}$ are extracted via:
\begin{equation}
    \hat{\mathbf{u}}_{l,m} = SHT[\mathbf{u}] = \int_{0}^{\pi} \int_{0}^{2\pi} \mathbf{u}(\theta, \phi) \overline{Y}_{l}^{m}(\theta, \phi) \sin \theta \, d\phi d\theta,
    \label{eq:SHT}
\end{equation}
where $\overline{Y}_{l}^{m}$ and $\sin \theta \, d\phi d\theta$ denote the conjugate spherical harmonic and the invariant measure on $\mathbb{S}^2$, respectively. The inverse transform (iSHT) then reconstructs the spatial field from the spectral domain up to a maximum degree $L$ of the spherical harmonic expansion:
\begin{equation}
    \mathbf{u}(\theta, \phi) = iSHT[\hat{\mathbf{u}}] = \sum_{l=0}^{L-1} \sum_{m=-l}^{l} \hat{\mathbf{u}}_{l,m} Y_{l}^{m}(\theta, \phi).
    \label{eq:iSHT}
\end{equation}
 Hereafter, we omit spectral coefficients with $m<0$ to eliminate redundancy, leveraging the spectral conjugate symmetry of real-valued fields.
\subsection{Why Spherical Harmonics Matter?}
\label{artifacts}
Standard neural operators typically model geophysical dynamics by treating $\mathbb{S}^2$ as Euclidean domains. This planar modeling fails to account for the sphere's intrinsic curvature and coordinate singularity. Specifically, their reliance on uniform feature extraction \textbf{violates the Nyquist bounds near topological singularities} (i.e., the geographical poles)~\cite{nyquist,shannon}, inevitably triggering spectral aliasing.

To investigate this phenomenon, we conduct a case study on Spherical Shallow Water Equation (SSWE) with several popular neural solvers. As illustrated in Figure~\ref{fig:comparison_six}(b)-(c), the geometric mismatch inevitably introduces severe radial artifacts near the poles. These artifacts are driven by spectral aliasing, originating at poles and compounding to overwhelm the valid prediction. Meanwhile, spherical neural operators adopt a different perspective by modeling geophysical dynamics directly in spectral coordinates of the Spherical Harmonic space. By transitioning the framework from the Euclidean plane to the spherical spectral domain, these models operate within a spectral representation that naturally respects the sphere geometry. As shown in Figure~\ref{fig:comparison_six}(d)-(f), this approach effectively eliminates polar artifacts. However, compared to our proposed \textsc{DiagNO}, the prediction results of SFNO and GSNO still suffer from energy dissipation, leading to blurred edges and fragmented structures. 
\section{Methodology}
\label{sec:method}
In this section, we model global dynamics as a heterogeneous process of cross-modal interactions. Firstly, we decompose the physical states, deriving a parameterized modal interaction kernel. Leveraging the unique diagonal structure of spectrum, we factorize the kernel along spectral diagonals to preserve geophysical regularities. Building upon this formulation, we design the \textsc{DiagNO} module to govern the modal interactions. Finally, by integrating these modules into a multi-resolution hierarchical architecture~\cite{U-NET_15}, our framework captures the dominant dynamics at varying spatial scales, facilitating a comprehensive mapping of Earth's dynamics.
\subsection{Formulation of Cross-Modal Interaction}
In this subsection, instead of relying on the rotation equivariance assumption that isolates modal evolution, we explicitly formulate cross-modal interactions to preserve the heterogeneity of Earth dynamics.  We first derive an interaction kernel through a direct modal decomposition. To ground this formulation in actual geophysical regularities, we factorize the kernel utilizing the unique diagonal structure of spherical spectrum, yielding a formulation that captures zonal and meridional dynamics.
\paragraph{Cross-Modal Kernel Parameterization.}
Driven by the need to explicitly parameterize the cross-modal interaction, we perform modal decomposition on the physical state fields. The Galerkin projection~\cite{galerkin_projection} formalizes the  decomposition by representing the physical states $\mathbf{u}(x), \mathbf{v}(x) \in \mathbb{R}^{C}$ as linear combinations of a set of orthogonal basis functions $\{\Psi_k(x)\}$ that span the functional space:
\begin{equation}
    \mathbf{u}(x) = \sum_{i} \hat{\mathbf{u}}_i \Psi_i(x), \quad \mathbf{v}(x) = \sum_{j} \hat{\mathbf{v}}_j \Psi_j(x),
\end{equation}
where $\hat{\mathbf{u}}_i, \hat{\mathbf{v}}_j \in \mathbb{C}^{C}$ denote the modal coefficients associated with the modes. By specifying these basis functions as Spherical Harmonics, the SHT (Equation \ref{eq:SHT}) derives the target modal coefficient as $\hat{\mathbf{v}}_{l,m} = \int_{\mathbb{S}^2} \mathbf{v}(x) \overline{Y}_{l}^{m}(x) dx$. Applying the integral operator in Equation \ref{eq:integral}, $\hat{\mathbf{v}}_{l,m}$ is given by:
\begin{equation}
    \hat{\mathbf{v}}_{l,m} = \int_{\mathbb{S}^2} \left[ \int_{\mathbb{S}^2} \boldsymbol{\kappa}(x, x')\mathbf{u}(x')dx' \right] \overline{Y}_{l}^{m}(x) dx.
\end{equation}
By substituting the iSHT representation of the input state $\mathbf{u}(x')$ from Equation \ref{eq:iSHT} and interchanging the order of integration and summation, the formulation becomes:
\begin{equation}
\begin{aligned}
    \hat{\mathbf{v}}_{l,m} &= \int_{\mathbb{S}^2} \left[ \int_{\mathbb{S}^2} \boldsymbol{\kappa}(x, x') \left( \sum_{l',m'} \hat{\mathbf{u}}_{l',m'} Y_{l'}^{m'}(x') \right) dx' \right] \overline{Y}_l^m(x) dx \\
    &= \sum_{l',m'} \underbrace{ \left( \iint_{\mathbb{S}^2 \times \mathbb{S}^2} \overline{Y}_l^m(x) \boldsymbol{\kappa}(x, x') Y_{l'}^{m'}(x') dx dx' \right) }_{\mathcal{W}_{(l,m),(l',m')}} \hat{\mathbf{u}}_{l',m'},
\end{aligned}
\end{equation}
where $\hat{\mathbf{u}}_{l',m'}$ are spectral modal coefficients of $\mathbf{u}(x')$. Through this transformation, we define the kernel of modal interaction $\mathcal{W}_{(l,m),(l',m')}$, which encapsulates interaction between input spectral mode $(l', m')$ and target spectral mode $(l, m)$. Therefore, the continuous spatial operator in Equation~\ref{eq:integral} is distilled into a discrete spectral representation:
\begin{equation}
    \hat{\mathbf{v}}_{l,m} = \sum_{l',m'} \mathcal{W}_{(l,m),(l',m')} \hat{\mathbf{u}}_{l',m'}.
    \label{eq:dense_w}
\end{equation}
In this formulation, $\mathcal{W}$ serves as a parameterized kernel that enables interactions across different modes. By parameterizing cross-modal interactions, our framework provides mathematical flexibility to preserve the heterogeneity of global dynamics.
\paragraph{Diagonal Kernel Factorization.}
While a dense interaction kernel $\mathcal{W}$ theoretically permits arbitrary interactions between any two spectral modes, this unconstrained formulation contradicts actual physical regularities. In geophysical fluid dynamics, the global evolution is primarily dominated by zonal patterns, such as global zonal climate belts~\cite{zonal}, whereas meridional processes act as auxiliary pathways for energy exchange~\cite{meridional_1} and oceanic heat transport~\cite{meridional_2}. 

We observe that the diagonal structure of Spherical Harmonic spectrum inherently characterizes the distribution of the orthogonal directions. Formally, the spectral modal coefficients $\hat{\mathbf{u}}_{l,m}$ aligned along the same diagonal direction share a constant degree-order offset $k$. We group these components by defining a diagonal index set $\mathcal{D}_k = \{ (l, m) \mid l - m + 1 = k \}$ for each offset, where $0 \le m \le l \le L-1$, and $k \in \{1, \dots, L\}$ indexes the $K = L$ diagonal sets. Specifically, modes within the same $\mathcal{D}_k$ (intra-diagonal) share a unified zonal pattern, where an increasing degree $l$ proportionally raises the longitudinal wavenumber to capture denser meridional oscillations. In contrast, modes across different sets $\mathcal{D}_k$ dictate distinct zonal patterns, where an increasing index $k$ proportionally raises the latitudinal wavenumber to reflect progressively denser zonal structures. Inspired by the diagonal structure of the Spherical Harmonic spectrum, we factorize the operator $\mathcal{W}$ into intra-diagonal and inter-diagonal operations to characterize modal interactions along the orthogonal directions:
\begin{equation}
    \mathcal{W} = \mathcal{W}_{\text{rec}} \mathcal{A}_{\text{attn}} \mathcal{W}_{\text{trans}},
\end{equation}
where $\mathcal{W}_{\text{trans}}$, $\mathcal{A}_{\text{attn}}$, and $\mathcal{W}_{\text{rec}}$ denote the \emph{intra-diagonal transformation operator}, the \emph{inter-diagonal evolution operator}, and the \emph{intra-diagonal reconstruction operator}, respectively.
\subsection{Design of \textsc{DiagNO} Module}
\label{subsec:DiagNO}
\begin{figure}
  \centering
  \includegraphics[width=\linewidth]{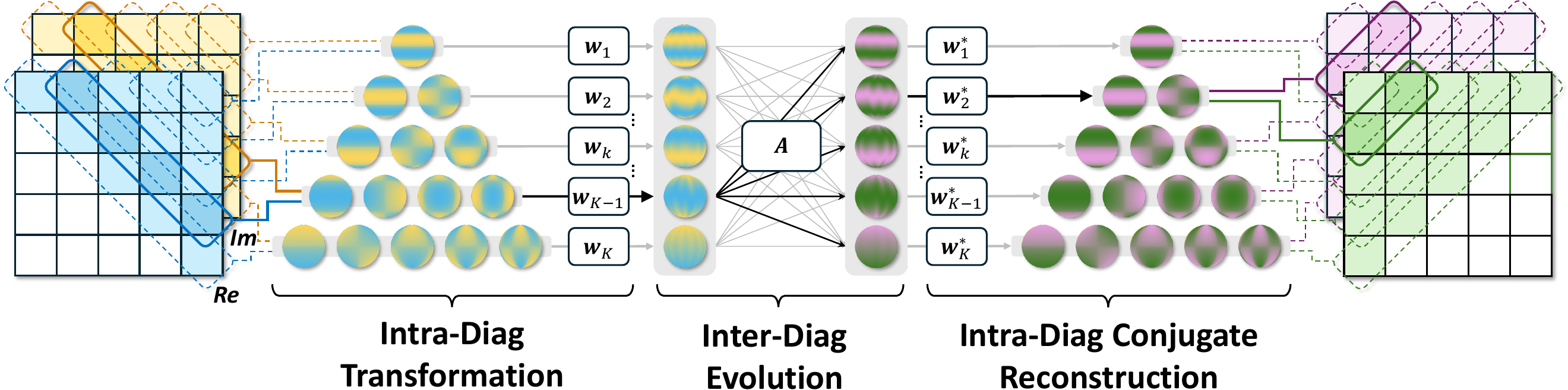} 
  \caption{\textbf{\textsc{DiagNO} module architecture.} Complex spectral coefficients are represented as triangular matrices and evolved through sequential intra-diagonal transformation, inter-diagonal evolution, and intra-diagonal conjugate reconstruction. Spheres illustrate spatial patterns of spectral modes.}
  \label{fig:DiagNO}
  \vspace{-12pt}
\end{figure}

Based on the factorized interaction, we design the \textsc{DiagNO} module to operate within the spherical spectral domain, capturing complex physical dynamics through three consecutive steps: intra-diagonal transformation, inter-diagonal evolution, and intra-diagonal conjugate reconstruction. The complete architectural design is illustrated in Figure \ref{fig:DiagNO}. Through these operations, the module achieves a complete  parameterized mapping of spatial physical states for Earth systems.
\paragraph{Intra-Diagonal Transformation.}
To capture the meridional interactions, we modulate the transformation among modes sharing the same zonal pattern. During this intra-diagonal transformation, we introduce a complex learnable aggregation tensor $\widetilde{\mathbf{w}}_k \in \mathbb{C}^{C \times |\mathcal{D}_k|}$ for each set $\mathcal{D}_k$. This tensor learns to quantify the relative contributions and dynamic redistributions among meridional oscillations. We apply a softmax normalization to the amplitude of $\widetilde{\mathbf{w}}_k$ over each $\mathcal{D}_k$ while retaining the phase as $\mathbf{w}_k = \text{Softmax}(|\widetilde{\mathbf{w}}_k|) \odot (\widetilde{\mathbf{w}}_k / |\widetilde{\mathbf{w}}_k|)$, where $\odot$ denotes the Hadamard product.  The intra-diagonal transformation operator $\mathcal{W}_{\text{trans}}$ synthesizes these meridional interactions, aggregating spectral modes sharing the identical zonal pattern into a unified aggregated zonal mode $\hat{\mathbf{z}}_k$:
\begin{equation}
    \hat{\mathbf{z}}_k = [\mathcal{W}_{\text{trans}} \hat{\mathbf{u}}]_k = \sum_{n=1}^{|\mathcal{D}_k|} \mathbf{w}_k^{(n)} \odot \hat{\mathbf{u}}_{\mathcal{D}_k}^{(n)},
\end{equation}
where $\mathbf{w}_k^{(n)} \in \mathbb{C}^C$ denotes the $n$-th column of the learnable aggregation weight matrix $\mathbf{w}_k$,  and $\hat{\mathbf{u}}_{\mathcal{D}_k}^{(n)} \in \mathbb{C}^C$ represents the $n$-th spectral mode coefficient within the diagonal set $\mathcal{D}_k$.

\paragraph{Inter-Diagonal Evolution.} 
To explicitly capture the inter-diagonal evolution (i.e., the interactions among different zonal patterns), we introduce a physically-grounded attention mechanism. Let $\mathbf{q}_k \in \mathbb{C}^C$ denote a learnable latent query refined by a self-attention layer. This query encodes the inter-zonal transfer tendency of aggregated zonal modes. We compute an interaction matrix $\mathbf{A}(\hat{\mathbf{u}}) \in \mathbb{R}^{K \times K}$ to quantify the coupling strength between $\mathbf{q}_k$ and $\hat{\mathbf{z}}_{k'}$. To reflect physical cross-correlation, the attention scores are evaluated using the real part of their complex inner product:
\begin{equation}
    A_{k,k'} = \frac{\exp \left( \frac{1}{\sqrt{C}} \operatorname{Re} \left[ \mathbf{q}_k^H \hat{\mathbf{z}}_{k'} \right] \right)}{\sum_{j=1}^{K} \exp \left( \frac{1}{\sqrt{C}} \operatorname{Re} \left[ \mathbf{q}_k^H \hat{\mathbf{z}}_j \right] \right)},
\end{equation}
where $(\cdot)^H$ denotes the conjugate transpose operator. Driven by these coupling weights, energy is dynamically redistributed across the inter-zonal spectrum. We define this process as the inter-diagonal evolution operator $\mathcal{A}_{\text{attn}}$, forming the evolved zonal modes $\tilde{\mathbf{z}}_k$:
\begin{equation}
    \tilde{\mathbf{z}}_k = [\mathcal{A}_{\text{attn}} \hat{\mathbf{z}}]_k = \sum_{k'=1}^{K} A_{k, k'} \hat{\mathbf{z}}_{k'}.
\end{equation}

\paragraph{Intra-Diagonal Conjugate Reconstruction.}
\label{para:conjugate}
To complete the modal interaction process, we map the evolved zonal modes back to the spectral domain to reconstruct the target state $\hat{\mathbf{v}}_{l,m}$:
\begin{equation}
    \hat{\mathbf{v}}_{l,m} = [\mathcal{W}_{\text{rec}} \tilde{\mathbf{z}}]_{l,m} = (\mathbf{w}_k^{(n)})^* \odot \tilde{\mathbf{z}}_k,
    \label{eq:rec}
\end{equation}
where $(\cdot)^*$ denotes the element-wise conjugate and $n = m + 1$. As formulated in Equation \ref{eq:rec}, rather than introducing an independent set of learnable parameters for reconstruction, we directly utilize the conjugate of the aggregation tensor $\mathbf{w}_k$. This design is fundamentally rooted in the Best Approximation Theorem~\cite{kreyszig1991introductory, brezis2011functional} within the context of Hilbert spaces. The proof is provided in the Appendix \ref{proof:best_approximation}. By composing the above three operators, the factorized spectral coupling process can be explicitly parameterized into its final expansion to obtain the target state $\hat{\mathbf{v}}_{l,m}$:
\begin{equation}
\begin{aligned}
    \hat{\mathbf{v}}_{l,m} &= \left[ (\mathcal{W}_{\text{rec}} \mathcal{A}_{\text{attn}} \mathcal{W}_{\text{trans}}) \hat{\mathbf{u}} \right]_{l,m}
    = \sum_{l',m'} \left[ (\mathbf{w}_k^{(m+1)})^* \odot A_{k, k'}(\hat{\mathbf{u}}) \mathbf{w}_{k'}^{(m'+1)} \right] \odot \hat{\mathbf{u}}_{l',m'}.
\end{aligned}
\end{equation}
Finally, the \textsc{DiagNO} module formally defines the end-to-end transition from the input spatial state $\mathbf{u}(x)$ to the target state $\mathbf{v}(x)$ as:
\begin{equation}
    \mathbf{v}(x) = iSHT \left[(\mathcal{W}_{\text{rec}} \mathcal{A}_{\text{attn}} \mathcal{W}_{\text{trans}}) SHT[\mathbf{u}(x)] \right].
\end{equation}



\section{Experiment}
We evaluate \textsc{DiagNO} through comprehensive experiments, addressing four key research questions: versatility (RQ1), utility (RQ2), interpretability (RQ3), and necessity (RQ4):

\textbf{RQ1:} Does \textsc{DiagNO} achieve consistent SOTA performance across diverse geophysical scenarios? \\
\textbf{RQ2:} Does \textsc{DiagNO} provide practical analytical value within professional geophysical domains? \\
\textbf{RQ3:} Does \textsc{DiagNO} offer mechanistic interpretability for scale-specific global dynamics? \\
\textbf{RQ4:} How do the diagonal designs and hyperparameter settings contribute to model effectiveness?
\subsection{Experiment Setup}
\label{experiment setup}
\textbf{Benchmarks and Datasets.}
To evaluate \textsc{DiagNO} across diverse physical scenarios, we utilize one simulated benchmark and two real-world Earth system reanalysis datasets (detailed in Appendix \ref{app:benchmarks}): \\
\textbullet~ \emph{Spherical Shallow Water Equations}: The SSWE~\cite{swe_1, swe_2} dataset simulates large-scale fluid dynamics on planetary surfaces. The ground-truth trajectories are randomly generated using a pseudo-spectral method across 1024 training, 256 validation, and 256 test samples. \\
\textbullet~ \emph{Sea Surface Forecasting}: Using GLORYS12~\cite{glorys12} reanalysis data, we characterize ocean dynamics $SSH$, $SST$, $U$, $V$ across training (2013--20), validation (21--22), and testing (23--24) periods. \\
\textbullet~ \emph{Atmospheric Forecasting}: We utilize $500$~hPa ERA5~\cite{ERA5} reanalysis to model $Z$, $U$, $V$ fields at $6$-hour resolution, following the same temporal partitioning as the sea surface dataset.

\textbf{Baselines.}
Baselines include standard neural operators (FNO~\cite{FNO_21}, LSM~\cite{LSM_23}, Transolver~\cite{Transolver_24}, MambaNO~\cite{MambaNO_24} and AMG~\cite{AMG_25}) and spherical neural operators (SFNO~\cite{SFNO_23} and GSNO~\cite{GSNO_26}), as well as domain-specific models: LangYa~\cite{LangYa_24} for sea surface and ClimaX~\cite{ClimaX_23} for atmospheric forecasting. The specific hyperparameter settings for all evaluated baselines are detailed in Table \ref{tab:model_configs}.

Further details on training/inference, evaluation metrics, and computational complexity are provided in Appendices \ref{training and inference}, \ref{metrics}, and \ref{Computational}, respectively. All reported metrics are averaged over multiple independent runs to ensure reliability.
\subsection{Overall Performance (RQ1)}
\begin{wraptable}{r}{0.49\textwidth} 
  \centering
  \footnotesize
  \renewcommand{\arraystretch}{1.0} 
  \setlength{\tabcolsep}{2pt} 
  \caption{\textbf{Performance comparison for SSWE.} Spherical neural operators generally outperform standard baselines. \textsc{DiagNO} achieves the best performance, reducing averaged error by 35.2\% while maintaining competitive memory efficiency.}
  \label{tab:swe}
  \resizebox{\linewidth}{!}{ 
  \begin{tabular}{lccccc}
    \toprule
    \multirow{3}{*}{MODEL} & \multirow{2}{*}{\#MEM} & \multicolumn{2}{c}{5 hours} & \multicolumn{2}{c}{10 hours} \\
    \cmidrule(lr){3-4} \cmidrule(lr){5-6}
    & & MAE $\downarrow$ & RMSE $\downarrow$ & MAE $\downarrow$ & RMSE $\downarrow$ \\
    & (GB) & ($\times 10^3$) & ($\times 10^3$) & ($\times 10^3$) & ($\times 10^3$) \\
    \midrule
    Swin~\cite{Swin_21}       & 2.49  & 0.2285 & 0.2878 & 0.2545 & 0.3212 \\
    FNO~\cite{FNO_21}        & 2.50   & 0.2702 & 0.3390 & 0.2653 & 0.3326 \\
    LSM~\cite{LSM_23}        & 3.92  & 0.1038 & 0.1583 & 0.1905 & 0.2502 \\
    Transolver~\cite{Transolver_24}   & 5.61  & 0.2897 & 0.3636 & 0.2871 & 0.3591 \\
    MambaNO~\cite{MambaNO_24}     & 12.14 & 0.1393 & 0.1893 & 0.5848 & 0.7438 \\
    AMG~\cite{AMG_25}        & 3.06  & 0.2784 & 0.3491 & 0.2675 & 0.3355 \\
    \midrule
    SFNO~\cite{SFNO_23}       & 2.30   & 0.1055 & 0.1322 & 0.1973 & 0.2481 \\
    GSNO~\cite{GSNO_26}       & 9.63  & \underline{0.0761} & \underline{0.0955} & \underline{0.1113} & \underline{0.1400} \\
    \midrule
    \textbf{\textsc{DiagNO}} & 4.26  & \textbf{0.0448} & \textbf{0.0563} & \textbf{0.0790} & \textbf{0.0985} \\
    \bottomrule
  \end{tabular}
  }
\end{wraptable}
\textbf{Spherical Shallow Water Equation.} \quad
The 5-hour and 10-hour SSWE forecasting results are summarized in Table \ref{tab:swe}, where spherical neural operators generally outperform standard baselines. Specifically, \textsc{DiagNO} achieves the lowest error across all metrics, significantly surpassing other models. While some standard operators suffer from early predictive stagnation and blurred states, \textsc{DiagNO} maintains steady error growth, reflecting its capacity to capture continuous physical evolution. Detailed visualizations are provided in the Appendix \ref{app:sswe}.

\begin{figure}[htbp]
  \vspace{-5pt}
  \centering
  \includegraphics[width=\linewidth]{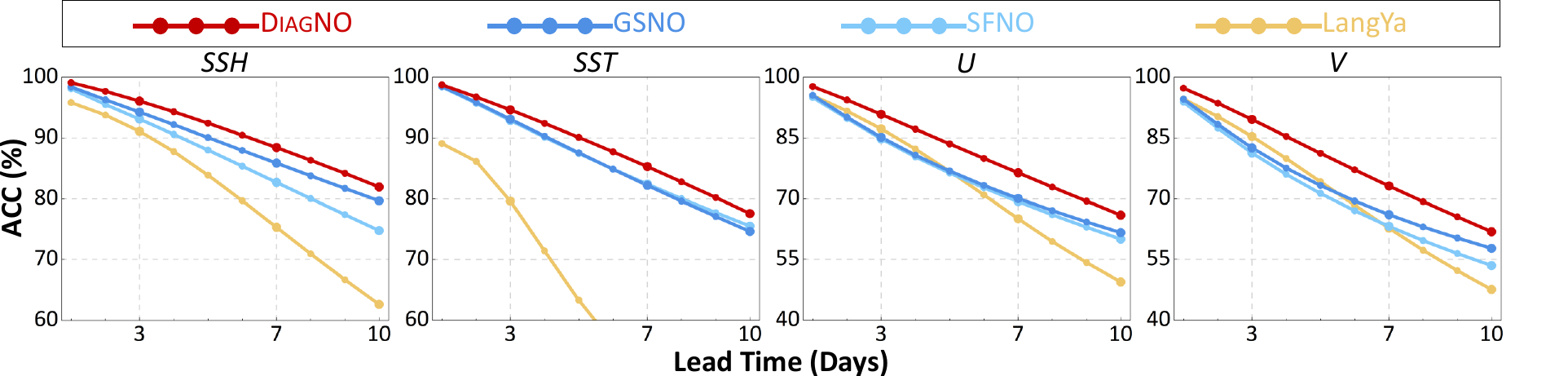} 
  \caption{\textbf{The ACC of ocean surface prediction (higher is better).} \textsc{DiagNO} exhibits superior stability across all lead times compared to spherical operators and the domain-specific LangYa model.}
  \label{fig:acc}
  \vspace{-12pt}
\end{figure}

\textbf{Sea Surface Forecasting.} \quad
In the sea surface forecasting experiment, we evaluate the daily autoregressive prediction performance over a 10-day lead time. Figure \ref{fig:acc} traces the continuous Anomaly Correlation Coefficient (ACC) of spherical neural operators and the marine forecasting model LangYa~\cite{LangYa_24}. Detailed comparisons involving all baselines are provided in Table \ref{app:ocean surface}, where \textsc{DiagNO} consistently achieves state-of-the-art performance. Notably, while spherical operators generally outpace LangYa, \textsc{DiagNO} exhibits the strongest temporal robustness. 
\paragraph{Atmospheric Forecasting.}
In atmospheric forecasting tasks, we evaluate the 72-hour and 120-hour predictive performance (Table \ref{tab:weather_prediction}). While the foundation model ClimaX~\cite{ClimaX_23} demonstrates commendable stability at extended lead times (120 hours), \textsc{DiagNO} consistently outperforms both ClimaX and existing spherical operators across all forecasting horizons. Notably, our model achieves a significant improvement in the ACC for the geopotential ($Z$) field, a critical metric for capturing large-scale circulations. This indicates that modeling atmospheric heterogeneity via our diagonal factorization provides an effective physical prior for atmospheric dynamics.
\begin{table}
  \vspace{-15pt}
  \centering
  \footnotesize
  \renewcommand{\arraystretch}{1.0}
  \setlength{\tabcolsep}{3pt} 
  \caption{\textbf{Performance comparison for ERA5 atmospheric forecasting.} \textsc{DiagNO} consistently outperforms all baselines across all lead times, with particularly strong improvements in $Z$ forecasting.}
  \label{tab:weather_prediction}
  \resizebox{\textwidth}{!}{ 
  \begin{tabular}{l cccc cc cccc cc}
    \toprule
    \multirow{3}{*}{MODEL} & \multicolumn{6}{c}{\textbf{72 Hours (12 steps)}} & \multicolumn{6}{c}{\textbf{120 Hours (20 steps)}} \\
    \cmidrule(lr){2-7} \cmidrule(lr){8-13}
    & MAE $\downarrow$ & RMSE $\downarrow$ & \multicolumn{4}{c}{ACC(\%) $\uparrow$} & MAE $\downarrow$ & RMSE $\downarrow$ & \multicolumn{4}{c}{ACC(\%) $\uparrow$} \\
    \cmidrule(lr){4-7} \cmidrule(lr){10-13}
    & ($\times10^3$) & ($\times10^3$) & $Z$ & $U$ & $V$ & Avg. & ($\times10^3$) & ($\times10^3$) & $Z$ & $U$ & $V$ & Avg. \\
    \midrule
    Swin~\cite{Swin_21}       & 0.2029 & 0.3014 & 68.97 & 59.98 & 55.36 & 61.44 & 0.3147 & 0.4627 & 38.53 & 33.21 & 27.17 & 32.97 \\
    FNO~\cite{FNO_21}        & 0.2974 & 0.4145 & 36.47 & 37.04 & 33.40 & 35.63 & 0.4710 & 0.6404 & 12.45 & 15.32 & 12.92 & 13.57 \\
    LSM~\cite{LSM_23}        & \underline{0.1697} & \underline{0.2503} & 73.64 & 62.36 & 58.28 & 64.76 & \underline{0.2448} & 0.3595 & 46.28 & 37.86 & 31.99 & 38.71 \\
    Transolver~\cite{Transolver_24} & 0.2956 & 0.4165 & 36.60 & 27.89 & 23.95 & 29.48 & 0.3918 & 0.5344 & 17.53 & 12.05 & 11.43 & 13.67 \\
    MambaNO~\cite{MambaNO_24}    & 0.2694 & 0.5448 & 68.57 & \underline{67.13} & \underline{65.12} & \underline{66.94} & 0.3869 & 0.6497 & 34.36 & 40.08 & 33.77 & 36.07 \\
    AMG~\cite{AMG_25}        & 0.1806 & 0.2619 & 71.62 & 57.19 & 54.66 & 61.16 & 0.2615 & 0.3765 & 40.58 & 31.82 & 27.16 & 33.18 \\
    \midrule
    SFNO~\cite{SFNO_23}       & 0.1973 & 0.2782 & 65.62 & 57.90 & 55.42 & 59.65 & 0.2594 & 0.3670 & 40.37 & 36.11 & 31.97 & 36.15 \\
    GSNO~\cite{GSNO_26}       & 0.1801 & 0.2629 & \underline{73.86} & 61.86 & 58.18 & 64.63 & 0.2600 & 0.3795 & \underline{47.12} & 38.26 & 31.38 & 38.92 \\
    \midrule
    ClimaX~\cite{ClimaX_23}     & 0.1885 & 0.2630 & 70.35 & 64.39 & 62.41 & 65.72 & 0.2574 & \underline{0.3576} & 45.88 & \underline{42.85} & \underline{38.86} & \underline{42.53} \\
    \midrule
    \textbf{\textsc{DiagNO}} & \textbf{0.1505} & \textbf{0.2344} & \textbf{81.18} & \textbf{67.87} & \textbf{65.14} & \textbf{71.40} & \textbf{0.2264} & \textbf{0.3549} & \textbf{56.72} & \textbf{44.93} & \textbf{39.88} & \textbf{47.18} \\
    \bottomrule
  \end{tabular}
  }
  \vspace{-12pt}
\end{table}
\begin{figure}
  \vspace{2pt}
  \centering
  \includegraphics[width=\linewidth]{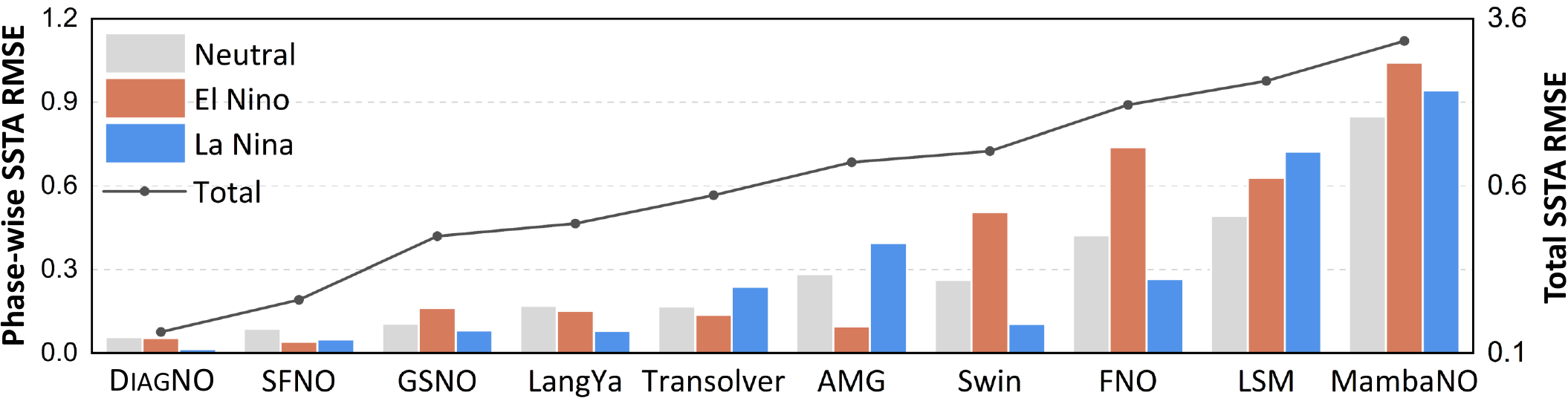} 
  \caption{\textbf{SSTA RMSE comparison.} Left axis: phase-wise SSTA RMSE (bars). Right axis: total SSTA RMSE (line) on a logarithmic scale for visual compactness. See Table \ref{tab:enso_rmse} for exact values.}
  \label{fig:enso}
  \vspace{-26pt}
\end{figure}
\subsection{Practical Analytical Value: Anomalies and Physical Consistency (RQ2)}
\begin{wrapfigure}{r}{0.45\textwidth}
  \centering
  \vspace{-20pt} 
\includegraphics[width=0.43\textwidth]{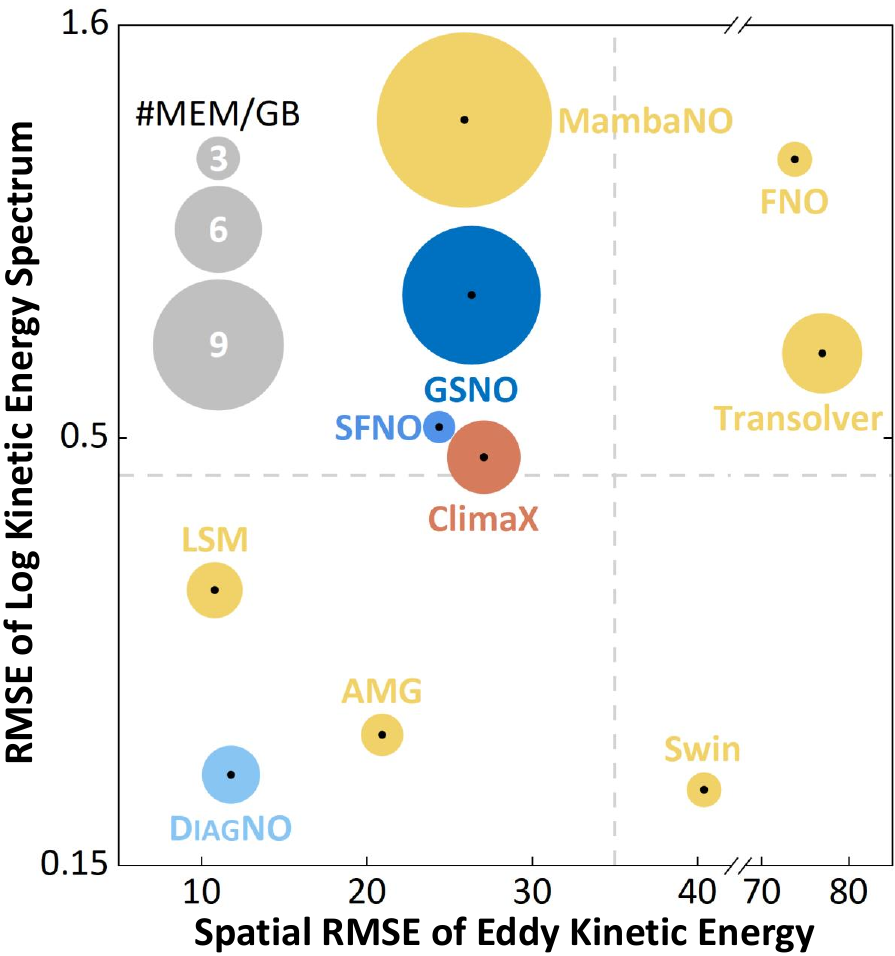} 
  \caption{\textbf{Physical consistency analysis.} X-axis: Spatial EKE RMSE. Y-axis: Log KES RMSE on a logarithmic scale. ClimaX shows competitive predictive performance second to \textsc{DiagNO}, yet falls considerably short in energy consistency.}
  \label{fig:scatter}
  \vspace{-20pt} 
\end{wrapfigure}
To verify practical analytical value, we extend our evaluation beyond general metrics to focus on two advanced dimensions: sensitivity to climatological anomalies and physical consistency in energetics.

\textbf{Climatological Anomaly.}\quad
We evaluate climate anomaly capturing via 7-day \textbf{Sea Surface Temperature Anomaly (SSTA)} forecasts in the Niño 3.4 region~\cite{ENSO_2} (Figure \ref{fig:enso}). Using $\pm 0.5^\circ\text{C}$ thresholds to classify ENSO phases~\cite{ENSO_1}, \textsc{DiagNO} significantly outperforms baselines in Neutral and La Niña conditions, while marginally trailing SFNO during extreme El Niño. This balanced performance achieves the lowest cumulative error across the entire ENSO cycle, confirming the practical utility of \textsc{DiagNO}'s forecasts in characterizing climatological anomalies.

\textbf{Physical Consistency: Energy Cascades.} \quad
 Based on the atmospheric forecasting results, we assess the physical consistency of the 120-hour wind predictions ($U, V$). Specifically, we analyze the \textbf{Kinetic Energy Spectrum (KES)}~\cite{KES_1,KES_2} and \textbf{Eddy Kinetic Energy (EKE)}~\cite{EKE} to quantify the intensity of transient weather systems and verify energy distribution across scales. As depicted in Figure \ref{fig:scatter}, \textsc{DiagNO} achieves the most accurate trade-off between maintaining spatial intensity and preserving the spectral distribution. By explicitly modeling the inherent heterogeneity of atmospheric systems via our diagonal factorization, the model effectively respects the dynamic energy cascades.
\subsection{Interpreting Dominant Dynamics: A Scale-Resolution Correspondence (RQ3)}
\begin{figure}
  \centering
  \begin{subfigure}[b]{0.13\textwidth}
    \centering
    \includegraphics[width=\textwidth, page=1]{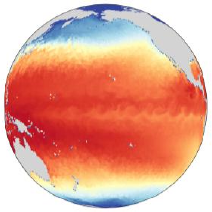}
    \caption{} 
  \end{subfigure}
  \hfill 
  \begin{subfigure}[b]{0.13\textwidth}
    \centering
    \includegraphics[width=\textwidth, page=2]{imgs/middle_result/middle_result.pdf}
    \caption{} 
  \end{subfigure}
  \hfill
  \begin{subfigure}[b]{0.13\textwidth}
    \centering
    \includegraphics[width=\textwidth, page=3]{imgs/middle_result/middle_result.pdf}
    \caption{} 
  \end{subfigure}
  \hfill
  \begin{subfigure}[b]{0.13\textwidth}
    \centering
    \includegraphics[width=\textwidth, page=4]{imgs/middle_result/middle_result.pdf}
    \caption{} 
  \end{subfigure}
  \hfill
  \begin{subfigure}[b]{0.13\textwidth}
    \centering
    \includegraphics[width=\textwidth, page=5]{imgs/middle_result/middle_result.pdf}
    \caption{} 
  \end{subfigure}
  \hfill
  \begin{subfigure}[b]{0.13\textwidth}
    \centering
    \includegraphics[width=\textwidth, page=6]{imgs/middle_result/middle_result.pdf}
    \caption{} 
  \end{subfigure}
  \caption{\textbf{Scale-specific dominant mode reconstructions.} \textbf{(a)} Actual $SST$ field. \textbf{(b-f)} iSHT representations reveal hierarchical extraction: \textbf{(b)} lowest-resolution interhemispheric asymmetry; \textbf{(c-d)} intermediate-resolution climate belts; \textbf{(e-f)} high-resolution fine asymmetric details.}
  \label{fig:middle_results} 
  \vspace{-15pt}
\end{figure}
To verify how \textsc{DiagNO} captures dominant dynamics across resolutions, we analyze its cross-attention matrices generated during Sea Surface Temperature ($SST$) forecasting, which quantify the coupling between aggregated zonal modes and latent queries. We select the dominant modes by identifying those with the maximum attention weights, representing the highest evolutionary tendency for energy redistribution. Reconstructing the $SST$ field from these modes via iSHT (Figure \ref{fig:middle_results}) reveals a scale-resolution correspondence and their distinct physical roles. Specifically, lower-resolution layers establish macroscopic foundational dynamics, such as interhemispheric asymmetry and planetary climate belts, whereas higher-resolution layers are responsible for high-frequency spatial waves and fine-grained asymmetric details. This explicit decoupling validates \textsc{DiagNO}'s ability to adaptively coordinate interpretable, scale-specific modal interactions for complex global forecasting.
\subsection{Ablation and Hyperparameter Analysis (RQ4)}
\label{ablation and hyperparameter} 
\begin{wrapfigure}{r}{0.48\textwidth} 
  \centering

  \captionof{table}{\textbf{Ablation study.} Evaluation of 10-day sea surface forecasting. \textsc{DiagNO} consistently outperforms $l$-dim and $m$-dim variants.}
  \label{ablation}
  \footnotesize
  \setlength{\tabcolsep}{3pt} 
  \begin{tabular}{l c c c c c c}
    \toprule
    \multirow{2}{*}{MODEL} & \multirow{2}{*}{MAE$\downarrow$} & \multirow{2}{*}{RMSE$\downarrow$} & \multicolumn{4}{c}{ACC(\%)$\uparrow$} \\
    \cmidrule(lr){4-7}
    & & & $SSH$ & $SST$ & $U$ & $V$ \\
    \midrule
    $m$-dim & \underline{0.1738} & \underline{0.2512} & \underline{80.97} & \underline{76.44} & \underline{65.04} & \underline{61.37} \\
    $l$-dim & 0.1939 & 0.2968 & 79.65 & 75.87 & 64.41 & 60.15 \\
    \textsc{DiagNO} & \textbf{0.1638} & \textbf{0.2387} & \textbf{81.93} & \textbf{77.57} & \textbf{65.97} & \textbf{61.89} \\
    \bottomrule
  \end{tabular}

  \vspace{15pt} 

  \includegraphics[width=0.45\textwidth]{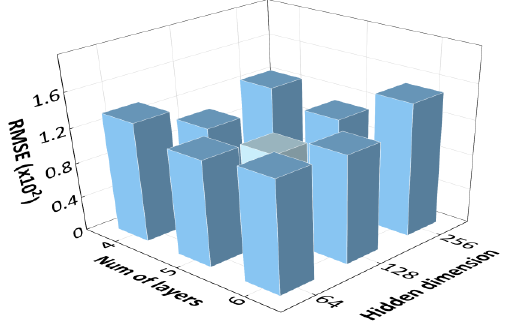}
  \captionof{figure}{\textbf{Hyperparameter analysis.} 10-hour prediction RMSE on SSWE. Our configuration (5 multi-resolution layers, 128 dim) is optimal.}
  \label{hyperparameter}
  
  \vspace{-10pt} 
\end{wrapfigure}

To further validate \textsc{DiagNO}, we conduct ablation studies by replacing diagonal interactions with standard $l$-dim or $m$-dim transformations. As shown in Table \ref{ablation}, \textsc{DiagNO} consistently outperforms these variants in 10-day sea surface forecasting, underscoring the necessity of diagonal spectral coupling. Additionally, our hyperparameter configurations are substantiated by a sensitivity analysis on the Spherical Shallow Water Equations (Figure \ref{hyperparameter}).
\section{Conclusions \& Limitations}
\label{contributions & limitations}
\textbf{Conclusions.} \quad
We propose \textsc{DiagNO}, which abandons the over-idealized rotation equivariance of previous spherical neural operators to explicitly model cross-modal interactions. By leveraging the unique diagonal structure of the Spherical Harmonic spectrum, \textsc{DiagNO} natively captures global-scale zonal and meridional dynamics, preserving the intrinsic heterogeneity of Earth systems. Extensive evaluations across atmospheric and oceanic scenarios demonstrate that \textsc{DiagNO} achieves state-of-the-art forecasting performance while exhibiting strong practical value through high physical consistency and accurate anomaly capture.

\textbf{Limitations and Future Work.} \quad While \textsc{DiagNO} effectively captures global heterogeneity, its reliance on SHTs mathematically necessitates complete global grids. Consequently, it is restricted when applied to localized spherical domains, which is a fundamental constraint shared by all spherical spectral-based models. Future work will extend our diagonal interaction paradigm to regional modeling. For instance, developing localized representations to independently model polar dynamics remains our next direction.

\newpage

\section*{Acknowledgments}

This work was supported in part by the National Natural Science Foundation of China under Grant No. T2421002 and No. 62272293, and the Postdoctoral Fellowship Program of CPSF under Grant No. GZB20250806. The authors declare no competing interests.

{
\small
\bibliographystyle{unsrtnat}
\bibliography{reference}
}

\medskip

\newpage
\appendix
\section{Theoretical Supplements}

\subsection{The Mathematical Origin of Isolated Modes in Spherical Neural Operators}
\label{proof:isolated_modes}

To understand the structural limitations of conventional rotation-equivariant Spherical Neural Operators (SNOs), we trace the mathematical evolution from a general integral operator to a strictly point-wise multiplication in the spectral domain. This derivation reveals how the strict rotation equivariance assumption leads to isolated spectral modal evolution in conventional spherical convolutions. We emphasize that this analysis applies to strictly rotation-equivariant linear spherical convolution operators, rather than all existing SNOs, as recent methods such as GSNO~\cite{GSNO_26} have explored more general non-equivariant formulations.

\textbf{1. From General Integral Operator to Spherical Convolution}\\
As defined in Equation \ref{eq:integral}, the general integral operator $\mathcal{K}$ applied to an input state $\mathbf{u}$ on the two-dimensional sphere $\mathbb{S}^2$ is formulated as:
\begin{equation}
\mathbf{v}(x) = (\mathcal{K}\mathbf{u})(x) = \int_{\mathbb{S}^2} \boldsymbol\kappa(x, x') \mathbf{u}(x') dx',
\label{eq:general_integral_app}
\end{equation}
where $x, x' \in \mathbb{S}^2$ are spatial coordinates, $\boldsymbol\kappa(x, x')$ is the continuous kernel matrix representing interactions between $x$ and $x'$, and $dx'$ represents the spherical integration measure.

Learning a dense, globally dependent kernel $\boldsymbol\kappa$ involves prohibitive computational complexity and allows full cross-talk between all spatial points. To mitigate this, conventional rotation-equivariant spherical convolutions introduce a strong geometric prior: \textbf{continuous rotation equivariance}~\cite{Spherical_cnns}. Under the $\text{SO}(3)$ equivariance assumption, the two-point kernel $\boldsymbol\kappa(x, x')$ is forced to rely strictly on relative rotation rather than absolute geographic coordinates. By fixing a reference point (typically the north pole $n \in \mathbb{S}^2$), this symmetric kernel is redefined as a single-argument localized filter $\tilde{\boldsymbol\kappa}(x) \equiv \boldsymbol\kappa(x, n)$. Thus, under this strict equivariance constraint, the general integral operator reduces to a continuous spherical convolution~\cite{Spherical_cnns}:
\begin{equation}
(\mathbf{u}*\tilde{\boldsymbol\kappa})(x) = \int_{\text{SO}(3)}\mathbf{u}(Rn)\tilde{\boldsymbol\kappa}(R^{-1}x)dR,
\label{eq:spherical_conv}
\end{equation}
where $n\in \mathbb{S}^{2}$ denotes the north pole, and $R$ is an element of the three-dimensional rotation group $\text{SO}(3)$ with Haar measure $dR$. \textit{This strict equivariance assumption is the first step toward modal isolation, as it imposes homogeneous spatial interaction rules.}

\textbf{2. Spherical Harmonic Transform (SHT)}\\
To evaluate Equation \ref{eq:spherical_conv}, the Spherical Harmonic Transform is employed to decompose the spatial functions into their harmonic coefficients:
\begin{equation}
\text{$SHT$}[\mathbf{u}](l, m) = \int_{\mathbb{S}^2} \mathbf{u}(x)\overline{Y_l^m (x)} dx.
\label{eq:sht}
\end{equation}
Applying the SHT to the spherical convolution, we substitute Equation \ref{eq:spherical_conv} into Equation \ref{eq:sht}:
\begin{equation}
\text{$SHT$}[(\mathbf{u}*\tilde{\boldsymbol\kappa})](l,m)=\int_{\mathbb{S}^{2}}\left(\int_{\text{SO}(3)}\mathbf{u}(Rn)\tilde{\boldsymbol\kappa}(R^{-1}x)dR\right)\overline{Y_{l}^{m}(x)}dx.
\label{eq:sht_conv}
\end{equation}
By interchanging the integral order and applying the variable substitution $x^{\prime}=R^{-1}x$ (with $dx=dx^{\prime}$ preserving the measure under rotation), the inner integral isolates the kernel:
\begin{equation}
\int_{\mathbb{S}^{2}}\tilde{\boldsymbol\kappa}(x^{\prime})\overline{Y_{l}^{m}(Rx^{\prime})}dx^{\prime}.
\label{eq:inner_integral}
\end{equation}

\textbf{3. The Spherical Convolution Theorem and Modal Isolation}\\
To evaluate Equation \ref{eq:inner_integral}, we rely on the well-established Spherical Convolution Theorem~\cite{Spherical_cnns}. Analytically, rotating a Spherical Harmonic redistributes its representation across different frequency orders. However, under the strict $\text{SO}(3)$ equivariance constraint, the resulting convolution operator admits a simplified spectral representation that prevents explicit coupling between distinct harmonic modes.

\textit{This is the mathematical origin of modal isolation in strictly rotation-equivariant spherical convolutions.} The convolution theorem eliminates explicit cross-modal coupling within the linear spectral operator. Consequently, the integral operator simplifies into a strictly element-wise multiplication in the spectral domain:
\begin{equation}
\text{$SHT$}[(\mathbf{u}*\tilde{\boldsymbol\kappa})](l,m)= 2\pi\sqrt{\frac{4\pi}{2l+1}}\cdot \text{$SHT$}[\tilde{\boldsymbol\kappa}](l,0)\cdot \text{$SHT$}[\mathbf{u}](l,m).
\label{eq:final_theorem}
\end{equation}

Equation \ref{eq:final_theorem} is the mathematical manifestation of \textbf{isolated modes} under strict rotation equivariance. It demonstrates that the update of any specific mode $(l,m)$ depends exclusively on the isotropic kernel component $\text{$SHT$}[\tilde{\boldsymbol\kappa}](l,0)$ and its corresponding input coefficient $\text{$SHT$}[\mathbf{u}](l,m)$. Crucially, this strictly point-wise formulation prohibits explicit information exchange between different harmonic modes within the linear spherical convolution operator.

While this isolated mode structure provides computational efficiency, it imposes an isotropic inductive bias. However, Earth's atmospheric and oceanic systems inherently break strict $\text{SO}(3)$ symmetry due to pronounced zonal and meridional global-scale patterns, such as the Coriolis effect and land-sea contrasts.

\textbf{4. Beyond Strict Rotation Equivariance: GSNO and \textsc{DiagNO}}\\
It is worth noting that the above derivation characterizes strictly rotation-equivariant spherical convolutions rather than all existing SNOs. Recent work such as GSNO~\cite{GSNO_26} has explored non-equivariant formulations through Green's functions to better accommodate spatially heterogeneous dynamics. However, relaxing rotation equivariance alone does not explicitly establish a structured mechanism for cross-modal interactions in the spherical spectral domain.

In contrast, \textsc{DiagNO} directly derives an explicit cross-modal interaction kernel from the general integral operator, enabling interactions between distinct spherical harmonic modes. More importantly, by exploiting the unique diagonal structure of the spherical harmonic spectrum, we efficiently factorize these interactions into intra- and inter-diagonal operations that capture zonal and meridional dynamics. Therefore, the key contribution of \textsc{DiagNO} lies in the explicit parameterization and physically informed diagonal factorization of cross-modal spectral interactions, providing a structured and efficient approach to modeling heterogeneous Earth dynamics.

\subsection{Theoretical Justification for Parameter Sharing}
\label{proof:best_approximation}

In Section \ref{subsec:DiagNO} (Equation \ref{eq:rec}), we proposed directly utilizing the element-wise complex conjugate of the aggregation weight $\mathbf{w}_k$ for the reconstruction stage rather than introducing a completely independent set of parameters. This architectural design is mathematically justified by the Best Approximation Theorem~\cite{kreyszig1991introductory, brezis2011functional} in Hilbert spaces. The following derivation explicitly proves why the adjoint operator provides the theoretically optimal inductive bias for this specific intra-diagonal transformation.

\textbf{1. Aggregation and Latent Subspace Definition}\\
Let $\mathcal{H}$ be the complex Hilbert space of the input spectral modes for a specific diagonal set $\mathcal{D}_k$, equipped with the standard inner product. Let the set $\{ \hat{\mathbf{u}}_{\mathcal{D}_k}^{(n)} \}_{n=1}^{|\mathcal{D}_k|}$ represent these input modes.

As formulated in Section \ref{subsec:DiagNO}, the intra-diagonal transformation acts as a linear operator $\mathcal{W}_{\text{trans}}:~ \mathcal{H} \to~ \mathcal{Z}$, which projects the input modes onto a latent subspace $\mathcal{Z}$ to form the aggregated zonal mode $\hat{\mathbf{z}}_k$. This operator is parameterized by the weights $\mathbf{w}_k$ and defined via a sum of Hadamard products:
\begin{equation}
    \hat{\mathbf{z}}_k = [\mathcal{W}_{\text{trans}} \hat{\mathbf{u}}]_k = \sum_{n=1}^{|\mathcal{D}_k|} \mathbf{w}_k^{(n)} \odot \hat{\mathbf{u}}_{\mathcal{D}_k}^{(n)}.
\label{eq:aggregation_mapping}
\end{equation}

\textbf{2. Decoupling Structural Projection from Dynamical Evolution}\\
In the \textsc{DiagNO} module, the forward pass consists of intra-diagonal transformation ($\mathcal{W}_{\text{trans}}$), inter-diagonal evolution ($\mathcal{A}_{\text{attn}}$), and intra-diagonal reconstruction ($\mathcal{W}_{\text{rec}}$). The full block to compute the target state is formulated as $\hat{\mathbf{v}} = \mathcal{W}_{\text{rec}} (\mathcal{A}_{\text{attn}} (\mathcal{W}_{\text{trans}} \hat{\mathbf{u}}))$.

To prevent the model from entangling the structural geometric projection with the dynamic physical evolution, we enforce a strict inductive bias: the transformation and reconstruction operators should form a nearly lossless geometric pair, leaving the burden of modeling spectral dynamics entirely to the evolution operator $\mathcal{A}_{\text{attn}}$. 

Therefore, we conceptually isolate the structural mapping by considering the identity evolution case (i.e., $\mathcal{A}_{\text{attn}} = \mathcal{I}$). Our structural objective is to find the optimal reconstruction operator $\mathcal{W}_{\text{rec}}$ that minimizes the information loss of the projection itself:
\begin{equation}
    \min_{\mathcal{W}_{\text{rec}}} \| \hat{\mathbf{u}} - \mathcal{W}_{\text{rec}} \mathcal{W}_{\text{trans}} \hat{\mathbf{u}} \|_{\mathcal{H}}^2.
\label{eq:minimization_objective}
\end{equation}

\textbf{3. Best Approximation and the Adjoint Operator}\\
According to the Best Approximation Theorem in Hilbert spaces, the unique operator that minimizes this projection distance is the orthogonal projection onto the subspace spanned by the forward mapping. Mathematically, the optimal reconstruction operator is \emph{strictly proportional to the adjoint operator of the forward transformation}, which is denoted as $\mathcal{W}_{\text{trans}}^\dagger$.

Given our specific forward operator defined in Equation \ref{eq:aggregation_mapping} (a summation over Hadamard products), deriving its rigorous adjoint operator mapping an evolved zonal mode $\tilde{\mathbf{z}}_k$ back to the original space $\mathcal{H}$ yields precisely the Hadamard product with the element-wise complex conjugate for each $n$-th component:
\begin{equation}
    [\mathcal{W}_{\text{trans}}^\dagger \tilde{\mathbf{z}}_k]^{(n)} = (\mathbf{w}_k^{(n)})^* \odot \tilde{\mathbf{z}}_k.
\label{eq:rigorous_adjoint}
\end{equation}
Thus, the exact orthogonal projection dictates that $\mathcal{W}_{opt} \propto \mathcal{W}_{\text{trans}}^\dagger$.

\textbf{4. Practical Implementation in Neural Operators}\\
While a mathematically exact orthogonal projection requires an additional scaling inverse operator (i.e., $(\mathcal{W}_{\text{trans}}\mathcal{W}_{\text{trans}}^\dagger)^{-1}$), maintaining explicit calculations of this inverse is computationally prohibitive. In deep neural networks, this uniform scaling effect is intrinsically absorbed by subsequent normalization layers or dynamically compensated during parameter optimization.

Consequently, the essential geometric property of the optimal reconstruction is purely governed by its directional component—the adjoint operator. By explicitly defining our reconstruction operator as Equation \ref{eq:rec} in the main text:
\begin{equation}
    [\mathcal{W}_{\text{rec}} \tilde{\mathbf{z}}]_{l,m} = (\mathbf{w}_k^{(m+1)})^* \odot \tilde{\mathbf{z}}_k,
\end{equation}
we force the network to strictly obey the exact geometric constraint of $\mathcal{W}_{\text{trans}}^\dagger$. This inherently minimizes the spectral reconstruction error and functions as a highly effective physical inductive bias, preventing the model from wasting capacity on searching an unconstrained parameter space.
\section{Implementation Details}
\subsection{Benchmark Datasets}
\label{app:benchmarks}
\textbf{Spherical Shallow Water Equations} \\
The Spherical Shallow Water Equations (SSWE) dataset~\cite{swe_1,swe_2} is generated using a custom pseudo-spectral solver implemented with the \href{https://github.com/NVIDIA/torch-harmonics}{\texttt{torch\_harmonics}} library~\cite{SFNO_23}. The solver discretizes the sphere using an equiangular (Clenshaw-Curtis) grid at a spatial resolution of $256 \times 512$. To ensure numerical precision, the underlying physical computations (including the evaluation of the Coriolis force and Laplacian operators) are performed in the spectral domain. The system state is advanced in time using a 3rd-order Adams-Bashforth (AB3) explicit integration scheme. To prevent spectral blocking and maintain long-term numerical stability, an implicit hyperdiffusion operator is applied to the spectral coefficients at each timestep. The dataset consists of trajectories initialized from both random spectral noise and the Galewsky jet instability condition. The final output fields comprise three variables: geopotential height $H$, zonal velocity $U$, and meridional velocity $V$.

\textbf{Ocean Science Dataset (GLORYS12)} \\
This dataset is derived from the GLORYS12V1 global ocean reanalysis~\cite{glorys12}, providing a comprehensive representation of ocean state variables. The data is processed at a spatial resolution of $384 \times 768$. The temporal resolution is set to 1 day, which is suitable for tracking the evolution of mesoscale to large-scale ocean currents and sea surface anomalies. In this work, four channels at the sea surface depth were selected: $SSH$ (Sea Surface Height), $SST$ (Sea Surface Temperature), $U$ (Zonal Velocity), and $V$ (Meridional Velocity).

\textbf{Atmospheric Science Dataset (ERA5)} \\
The ERA5 dataset~\cite{ERA5} is a high-resolution global atmospheric reanalysis produced by ECMWF. It serves as a benchmark for evaluating the model's performance in modeling complex atmospheric dynamics and weather forecasting. The atmospheric fields are sampled at a spatial resolution of $256 \times 512$. The temporal resolution is 6 hours, capturing the standard synoptic-scale variations essential for extended-range weather prediction.  In this work, three atmospheric channels at the 500 hPa pressure level were selected: $Z$ (Geopotential), $U$ (Zonal Wind), and $V$ (Meridional Wind).
\subsection{Baselines and Hyperparameters}
\label{baselines}
\begin{table}[htbp]
  \centering
  \caption{Hyper-parameter configurations of different baseline models.}
  \label{tab:model_configs}
  \begin{tabular}{lccc}
    \toprule
    MODEL & Year & Hidden Dim & Blocks \\
    \midrule
    Swin~\cite{Swin_21}         & 2021  & 128  & 4 \\
    FNO~\cite{FNO_21}          & 2021  & 128  & 4 \\
    LSM~\cite{LSM_23}          & 2023  & 128  & 5 \\
    Transolver~\cite{Transolver_24}   & 2024  & 128  & 4 \\
    MambaNO~\cite{MambaNO_24}      & 2024  & 128  & 8 \\
    AMG~\cite{AMG_25}          & 2025  & 128  & 15 \\
    \midrule
    SFNO~\cite{SFNO_23}         & 2023  & 128  & 8 \\
    GSNO~\cite{GSNO_26}         & 2026  & 128  & 5 \\
    \midrule
    ClimaX~\cite{ClimaX_23}       & 2023  & 1024 & 8 \\
    LangYa~\cite{LangYa_24}       & 2024  & 128  & 8 \\
    \midrule
    \textbf{\textsc{DiagNO}}    & ---   & 128  & 5 \\
    \bottomrule
  \end{tabular}
\end{table}
The key hyper-parameters of baselines are summarized in Table~\ref{tab:model_configs}. For ClimaX, we leverage the foundation model pre-trained at a $1.40625^{\circ}$ resolution ($128 \times 256$) with a patch size of 4. To adapt this architecture to our higher $256 \times 512$ resolution of atmospheric forecasting, we scale the patch size to 8. This adjustment preserves the original sequence length of 2,048 tokens, enabling the direct reuse of pre-trained spatial position embeddings while maintaining the 1024-dimensional hidden state and 8-block depth. This configuration allows the model to effectively ingest high-resolution atmospheric data without sacrificing the learned spatial priors from its global pre-training.
\subsection{Training and Inference Implementation}
\label{training and inference}
\paragraph{Training.}
Most baseline models follow a training paradigm consisting of 100 epochs of one-step pre-training and 100 epochs of two-step autoregressive fine-tuning. Specifically, the one-step pre-training maps the system state to the next time step, serving to rapidly capture fundamental physical dynamics and provide a stable parameter initialization. Subsequently, the two-step autoregressive fine-tuning uses the model's own predictions as inputs for consecutive steps. This stage is designed to mitigate error accumulation and enhance the temporal coherence of long-term rollouts. For the pre-training stage, the Adam optimizer is employed with a learning rate of $10^{-3}$ for the SSWE task, whereas a rate of $10^{-4}$ is utilized for atmospheric and sea surface prediction tasks. The fine-tuning learning rate for all three benchmarks is set to $10^{-5}$.

Notably, we implement a weight transfer from the pre-trained model for ClimaX. The entire Transformer backbone is retained to utilize the learned global atmospheric dynamics. We perform targeted extraction of weights from the projection layers and channel embeddings corresponding to the indices of the target variables ($Z$, $U$, and $V$ at 500 hPa pressure level). Due to this extensive parameter initialization, ClimaX requires only 20 epochs of single-step fine-tuning. Regarding LangYa, its hierarchical Swin Transformer architecture prevents the direct extraction of channel-specific parameters from existing pre-trained ocean models. To ensure a fair comparison, we utilize its original architectural design but adhere to the standard training strategy of 100 epochs for both pre-training and fine-tuning.

\paragraph{Inference.}
Consistent with our fine-tuning phase, multi-step inference follows an autoregressive strategy. That is, the model recursively utilizes its output at step $t$ as the input for step $t+1$, enabling continuous forecasting over extended periods.

\paragraph{Computational Resources.}
All models, including \textsc{DiagNO} and the evaluated baselines, are trained and evaluated on a system equipped with an Intel(R) Xeon(R) Gold 5117 @ 2.0GHz CPU (up to 24 cores). All deep learning processes, including the autoregressive rollouts and physical metric evaluations, are accelerated using a combination of NVIDIA GeForce RTX 4090 (24GB memory) and NVIDIA A800 (80GB memory) GPUs.

\subsection{Evaluation Metrics}
\label{metrics}
\paragraph{Weighted $L^2$ Loss.} 
For the atmospheric and oceanic fields sampled on an equiangular $N_\theta \times N_\phi$ grid, we utilize area-weighted loss function to account for the spherical geometry. The Weighted $L^2$ Loss measures the aggregate deviation between the predicted field $f$ and the ground truth $g$. This metric is applied to both atmospheric and sea surface variables to ensure the evaluation reflects the global physical distribution. The loss is defined as:
\begin{equation}
\mathcal{L}_{L^2, w} = \sqrt{\sum_{i=1}^{N_\theta} \sum_{j=1}^{N_\phi} W_{i,j} (f_{i,j} - g_{i,j})^2}
\end{equation}
where $f_{i,j}$ and $g_{i,j}$ represent the predicted and ground-truth values at latitude index $i$ and longitude index $j$, respectively. The normalized quadrature weight $W_{i,j}$ for each grid cell is defined as:
\begin{equation}
W_{i,j} = \frac{w_i}{N_\phi \sum_{k=1}^{N_\theta} w_k}
\end{equation}
where $w_i$ are the latitudinal weights derived from the Clenshaw-Curtis quadrature rule~\cite{SFNO_23} over the interval $\cos(\theta) \in [-1, 1]$. This normalization ensures $\sum_{i,j} W_{i,j} = 1$, effectively compensating for the convergence of meridians at high latitudes.

\paragraph{Anomaly Correlation Coefficient (ACC).}
To evaluate the model's skill in capturing the spatial morphology and phase alignment of geophysical fluctuations, we employ the Anomaly Correlation Coefficient (ACC). It calculates the weighted correlation between the predicted and observed anomalies—defined as the deviations from the long-term climatological mean $c$:
\begin{equation}
ACC = \frac{\sum_{i,j} W_{i,j} (f_{i,j} - c_{i,j})(g_{i,j} - c_{i,j})}{\sqrt{\left[ \sum_{i,j} W_{i,j} (f_{i,j} - c_{i,j})^2 \right] \left[ \sum_{i,j} W_{i,j} (g_{i,j} - c_{i,j})^2 \right]}}
\end{equation}
The ACC utilizes the same quadrature weights $W_{i,j}$ as defined above. An ACC value of 1 indicates perfect alignment of the anomaly patterns.

\paragraph{Kinetic Energy Spectrum (KES) and Log KES RMSE.}
To evaluate the restoration of physical structures across scales, we compute the isotropic Kinetic Energy Spectrum (KES)~\cite{KES_1,KES_2}. Given the zonal ($u$) and meridional ($v$) velocity components, the 2D power spectrum $P(k_i, k_j)$ is derived from their 2D discrete Fourier transforms:
\begin{equation}
P(k_i, k_j) = \frac{1}{2} \left( |\mathcal{F}(u)|^2 + |\mathcal{F}(v)|^2 \right)
\end{equation}
where $\mathcal{F}(\cdot)$ denotes the 2D Fast Fourier Transform (FFT). The 1D isotropic spectrum $E(k)$ is obtained by radial averaging for each scalar wavenumber $k = \sqrt{k_i^2 + k_j^2}$. To quantify the discrepancy between the predicted spectrum $E_f(k)$ and the ground-truth spectrum $E_g(k)$, we compute the Log KES RMSE:
\begin{equation}
\text{RMSE}_{\text{Log KES}} = \sqrt{\frac{1}{k_{max}} \sum_{k=1}^{k_{max}} \left[ \log_{10} E_f(k) - \log_{10} E_g(k) \right]^2}
\end{equation}
where $k_{max} = \min(N_\theta, N_\phi) / 2$. This metric provides a scale-invariant assessment of spectral accuracy.

\paragraph{Eddy Kinetic Energy (EKE) Analysis.} 
To evaluate the energy associated with transient weather systems, the Eddy Kinetic Energy (EKE)~\cite{EKE} is computed as the kinetic energy of velocity perturbations relative to the climatological mean. For each grid point $(i, j)$ at time $t$, the EKE is defined as:
\begin{equation}
EKE_{i,j,t} = \frac{1}{2} \left[ (u_{i,j,t} - c_{u,i,j})^2 + (v_{i,j,t} - c_{v,i,j})^2 \right]
\end{equation}
where $c_u, c_v$ denote the climatological means of $u$ and $v$, respectively. To quantify the spatial accuracy of the predicted EKE field ($\widehat{EKE}$) compared to the ground truth ($EKE$), we employ the area-weighted RMSE:
\begin{equation}
RMSE_{EKE} = \sqrt{\sum_{i=1}^{N_\theta} \sum_{j=1}^{N_\phi} W_{i,j} (\widehat{EKE}_{i,j} - EKE_{i,j})^2}
\end{equation}
This metric evaluates the model's performance in representing the intensity and spatial distribution of atmospheric eddies.
\subsection{Computational Complexity and Efficiency Analysis}
\label{Computational}
\paragraph{Computational Complexity.}
To rigorously analyze the computational footprint of the \textsc{DiagNO} module, we first define the key variables related to the spatial and spectral dimensions:
\begin{itemize}
    \item $N = N_\theta \times N_\phi$: The total number of spatial grid nodes, where $N_\theta$ and $N_\phi$ represent the number of latitude and longitude grids, respectively.
    \item $C$: The number of hidden feature channels.
    \item $L_{max}$ and $M_{max}$: The truncation degrees of the Spherical Harmonic spectrum. To satisfy the Nyquist-Shannon sampling theorem~\cite{nyquist,shannon}, they are proportional to the spatial resolution, yielding $L_{max} \propto \sqrt{N}$ and $M_{max} \propto \sqrt{N}$.
    \item $K$: The total number of extracted diagonals, where $K = L_{max} \propto \sqrt{N}$.
    \item $|\mathcal{D}_k|$: The number of spectral modes contained within the $k$-th diagonal. The sum of all modes across all diagonals equals the total number of spectral modes, resulting in $\sum_{k} |\mathcal{D}_k| \approx L_{max} \times M_{max} \propto N$.
\end{itemize}

Based on these definitions, Table \ref{tab:complexity} summarizes the theoretical mathematical complexity of each operation within the \textsc{DiagNO} module. Note that mathematically, the intra-diagonal transformation and reconstruction only perform element-wise operations and accumulations exclusively on valid spectral modes (totaling $L_{max} \times M_{max}$ elements). Therefore, their theoretical complexity is tightly bounded by $\mathcal{O}(C \cdot N)$. Consequently, the interaction mechanism circumvents the typical $\mathcal{O}(N^2)$ bottleneck of global attention, ensuring that the overall complexity of the module remains strictly dominated by the unavoidable Spherical Harmonic Transforms ($\mathcal{O}(C \cdot N^{1.5})$).

\begin{table}[h]
\centering
\caption{Computational complexity of the \textsc{DiagNO} module. The complexity is measured with respect to the total number of spatial grid nodes $N$ and the hidden channel dimension $C$.}
\label{tab:complexity}
\renewcommand{\arraystretch}{1.2}
\begin{tabular}{@{}lc@{}}
\toprule
\textbf{Module / Process} & \textbf{Computational Complexity} \\ \midrule
Spherical Harmonic Transform (SHT) & $\mathcal{O}(C \cdot N^{1.5})$ \\
Intra-Diagonal Transformation & $\mathcal{O}(C \cdot N)$ \\
Inter-Diagonal Evolution & $\mathcal{O}(C \cdot N)$ \\
Intra-Diagonal Conjugate Reconstruction & $\mathcal{O}(C \cdot N)$ \\
Inverse SHT (iSHT) & $\mathcal{O}(C \cdot N^{1.5})$ \\ \midrule
\textbf{Overall \textsc{DiagNO} Module} & $\mathcal{O}(C \cdot N^{1.5})$ \\ \bottomrule
\end{tabular}
\end{table}
\begin{table}
\centering
\small 
\caption{Peak memory consumption (\#MEM) of different models across three benchmark datasets.}
\label{tab:memory_consumption}
\renewcommand{\arraystretch}{0.9} 
\begin{tabular}{@{}lccc@{}}
\toprule
\multirow{2}{*}{MODEL} & \multicolumn{3}{c}{\#MEM (GB)} \\ \cmidrule(l){2-4} 
                       & SSWE  & \makecell{Sea Surface \\ Forecasting} & \makecell{Atmospheric \\ Forecasting} \\ \midrule
Swin                & 2.49  & 4.50                                  & 2.41                                  \\
FNO                 & 2.50  & 3.83                                  & 2.40                                  \\
LSM                 & 3.92  & 6.78                                  & 3.84                                  \\
Transolver          & 5.61  & 12.28                                 & 5.54                                  \\
MambaNO             & 12.14 & 26.59                                 & 12.06                                 \\
AMG                 & 3.06  & 5.73                                  & 2.94                                  \\ \midrule
SFNO                & 2.30  & 4.55                                  & 2.22                                  \\
GSNO                & 9.63  & 19.42                                 & 9.55                                  \\ \midrule
LangYa                 & -     & 1.09                                  & -                                     \\
ClimaX                 & -     & -                                     & 5.06                                  \\ \midrule
\textsc{DiagNO}        & 4.26  & 7.23                                  & 3.99                                  \\ \bottomrule
\end{tabular}
\end{table}
\paragraph{Computational Overhead.}
Table \ref{tab:memory_consumption} details the peak memory consumption of different models across each benchmark with a batch size of 1.
\section{Experiment Results}
\subsection{Spherical Shallow Water Equation}
\label{app:sswe}
\begin{figure}[htbp]
  \centering
  
  \begin{subfigure}{\linewidth}
    \centering
    \includegraphics[width=\linewidth]{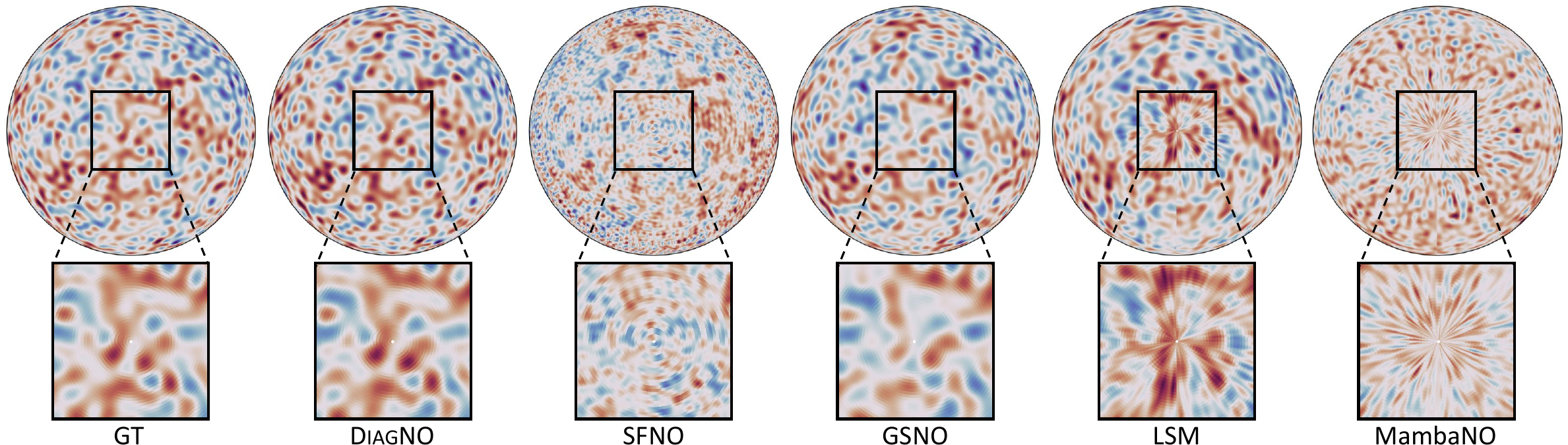}
    \caption{Forecasting results for the $H$ channel.}
    \label{fig:SSWE_h}
  \end{subfigure}
  
  \vspace{0.5cm} 

  \begin{subfigure}{\linewidth}
    \centering
    \includegraphics[width=\linewidth]{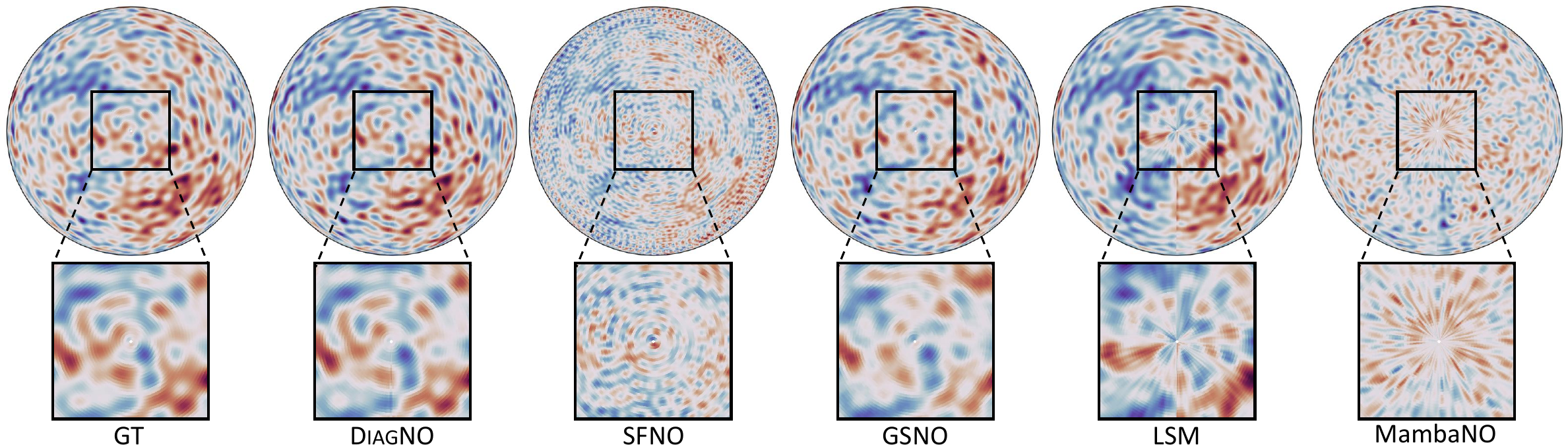}
    \caption{Forecasting results for the $V$ channel.}
    \label{fig:SSWE_v}
  \end{subfigure}

  \caption{Comparison of forecasting results for SSWE from a north pole perspective.}
  \label{fig:SSWE_total}
\end{figure}

While Figure~\ref{fig:comparison_six} presents the forecasting results for the zonal velocity $U$ in the Spherical Shallow Water Equations (SSWE), Figures~\ref{fig:SSWE_h} and \ref{fig:SSWE_v} provide corresponding visualizations of the height field $H$ and meridional velocity $V$ at the 10th prediction step for the six top-performing models. All these visualizations are presented from a North Pole perspective to highlight the dynamics in the polar region. As observed in these visualizations, our proposed \textsc{DiagNO} remains the most consistent with the ground truth, maintaining sharp, well-defined prediction patterns and clear boundaries. In contrast, although other spherical neural operators successfully respect the sphere’s underlying geometry and eliminate the coordinate singularities, they still exhibit varying degrees of energy dissipation, which manifests as blurred edges and fragmented structural details. Meanwhile, standard neural operators that rely on a planar Euclidean representation continue to suffer from prominent radial artifacts originating at the poles. These artifacts are a direct consequence of the aforementioned geometric mismatch and the violation of local Nyquist bounds, where spectral aliasing compounds over time to overwhelm the physical signal.
\subsection{Ocean Surface Forecasting}
\begin{table}[htbp]
  \centering
  \footnotesize 
  \caption{The overall performance comparison for ocean surface prediction.}
  \label{app:ocean surface}
  \begin{tabular}{lccccccc}
    \toprule
    \multirow{2}{*}{MODEL} & \multirow{2}{*}{MAE $\downarrow$} & \multirow{2}{*}{RMSE $\downarrow$} & \multicolumn{5}{c}{ACC(\%) $\uparrow$} \\
    \cmidrule(lr){4-8}
    & & & $SSH$ & $SST$ & $U$ & $V$ & Average \\
    \midrule
    \multicolumn{8}{c}{\textbf{3 Days}} \\
    \midrule
    Swin      & 0.2440 & 0.5262 & 52.12 & 26.20 & 45.15 & 37.50 & 40.24 \\
    FNO      & 0.2359 & 0.3173 & 80.09 & 56.90 & 80.64 & 76.81 & 73.61 \\
    LSM       & 0.1335 & 0.1758 & 89.98 & 83.78 & 90.15 & 88.85 & 88.19 \\
    Transolver   & 0.0972 & 0.1366 & \underline{94.62} & 92.05 & \underline{90.49} & \underline{89.04} & \underline{91.55} \\
    MambaNO     & 0.1314 & 0.1821 & 94.29 & 81.35 & 89.13 & 87.54 & 88.08 \\
    AMG      & 0.1632 & 0.3192 & 85.83 & 56.06 & 84.38 & 81.62 & 76.97 \\
    \midrule
    SFNO      & \underline{0.0854} & 0.1390 & 93.09 & 92.88 & 84.80 & 81.30 & 88.02 \\
    GSNO        & 0.0862 & \underline{0.1357} & 94.25 & \underline{93.09} & 85.16 & 82.57 & 88.77 \\
    \midrule
    LangYa      & 0.1425 & 0.2049 & 91.09 & 79.60 & 87.26 & 85.38 & 85.83 \\
    \midrule
    \textsc{DiagNO}    & \textbf{0.0751} & \textbf{0.1162} & \textbf{96.06} & \textbf{94.62} & \textbf{90.81} & \textbf{89.56} & \textbf{92.76} \\
    
    \midrule
    \multicolumn{8}{c}{\textbf{7 Days}} \\
    \midrule
    Swin      & 0.3446 & 0.6603 & 37.49 & 17.59 & 32.80 & 23.97 & 27.96 \\
    FNO       & 0.5057 & 0.6771 & 43.67 & 19.53 & 56.09 & 52.47 & 42.94 \\
    LSM       & 0.2606 & 0.3557 & 73.82 & 53.15 & 75.27 & 72.01 & 68.56 \\
    Transolver   & 0.1988 & 0.2650 & 85.78 & 71.85 & \underline{76.10} & \underline{72.06} & \underline{76.45} \\
    MambaNO     & 0.2498 & 0.3468 & 81.03 & 42.39 & 71.59 & 69.12 & 66.03 \\
    AMG      & 0.2791 & 0.6436 & 68.47 & 28.07 & 64.18 & 59.08 & 54.95 \\
    \midrule
    SFNO      & \underline{0.1382} & 0.2138 & 82.69 & \underline{82.41} & 69.30 & 63.21 & 74.40 \\
    GSNO      & 0.1411 & \underline{0.2115} & \underline{85.86} & 82.20 & 70.08 & 66.08 & 76.06 \\
    \midrule
    LangYa      & 0.2566 & 0.4023 & 75.30 & 50.33 & 65.08 & 62.75 & 63.37 \\
    \midrule
    \textsc{DiagNO}    & \textbf{0.1294} & \textbf{0.1921} & \textbf{88.42} & \textbf{85.29} & \textbf{76.42} & \textbf{73.15} & \textbf{80.82} \\
    
    \midrule
    \multicolumn{8}{c}{\textbf{10 Days}} \\
    \midrule
    Swin      & 0.4036 & 0.7082 & 30.78 & 14.97 & 27.61 & 20.33 & 23.42 \\
    FNO       & 0.9101 & 1.2212 & 25.71 & 8.33 & 35.87 & 35.24 & 26.29 \\
    LSM      & 0.3377 & 0.4630 & 62.96 & 38.74 & 65.26 & \underline{61.16} & 57.03 \\
    Transolver   & 0.2378 & 0.3194 & 77.04 & 62.34 & \underline{62.32} & 57.19 & 64.72 \\
    MambaNO      & 0.3303 & 0.4523 & 70.13 & 24.21 & 58.92 & 56.86 & 52.53 \\
    AMG       & 0.3522 & 0.7808 & 59.04 & 22.85 & 52.33 & 47.47 & 45.42 \\
    \midrule
    SFNO      & \underline{0.1682} & 0.2539 & 74.77 & \underline{75.51} & 60.05 & 53.51 & 65.96 \\
    GSNO      & 0.1726 & \underline{0.2533} & \underline{79.66} & 74.62 & 61.65 & 57.76 & \underline{68.43} \\
    \midrule
    LangYa       & 0.3407 & 0.5434 & 62.63 & 37.88 & 49.49 & 47.64 & 49.41 \\
    \midrule
    \textsc{DiagNO}    & \textbf{0.1638} & \textbf{0.2387} & \textbf{81.93} & \textbf{77.57} & \textbf{65.97} & \textbf{61.89} & \textbf{71.84} \\
    \bottomrule
  \end{tabular}
\end{table}
Table \ref{app:ocean surface} details the overall performance comparison for ocean surface prediction across 3-day, 7-day, and 10-day lead times. As shown in the results, spherical neural operators (SFNO, GSNO, and our proposed \textsc{DiagNO}) generally exhibit superior and more robust performance compared to other standard baseline models (such as Swin, FNO, and MambaNO). This general advantage highlights the critical importance of respecting spherical geometry when modeling global ocean dynamics. However, among these spherical architectures, \textsc{DiagNO} consistently achieves the best results, recording the lowest MAE and RMSE, alongside the highest Average ACC across all forecasting horizons. The performance gap becomes particularly pronounced at extended lead times. This sustained superiority validates our core architectural design. While conventional spherical operators such as SFNO rely on rotation-equivariant spectral formulations, and GSNO introduces non-equivariant extensions
through Green's functions, \textsc{DiagNO} follows a distinct route by explicitly parameterizing cross-modal spectral interactions. By doing so, our model is far better equipped to capture the highly heterogeneous dynamics inherent to complex ocean variables such as $SSH$ and $SST$, ultimately leading to more accurate and stable long-term rollouts.
\begin{figure}
    \centering
    \begin{subfigure}{\textwidth}
        \centering
        \includegraphics[width=\linewidth, page=1]{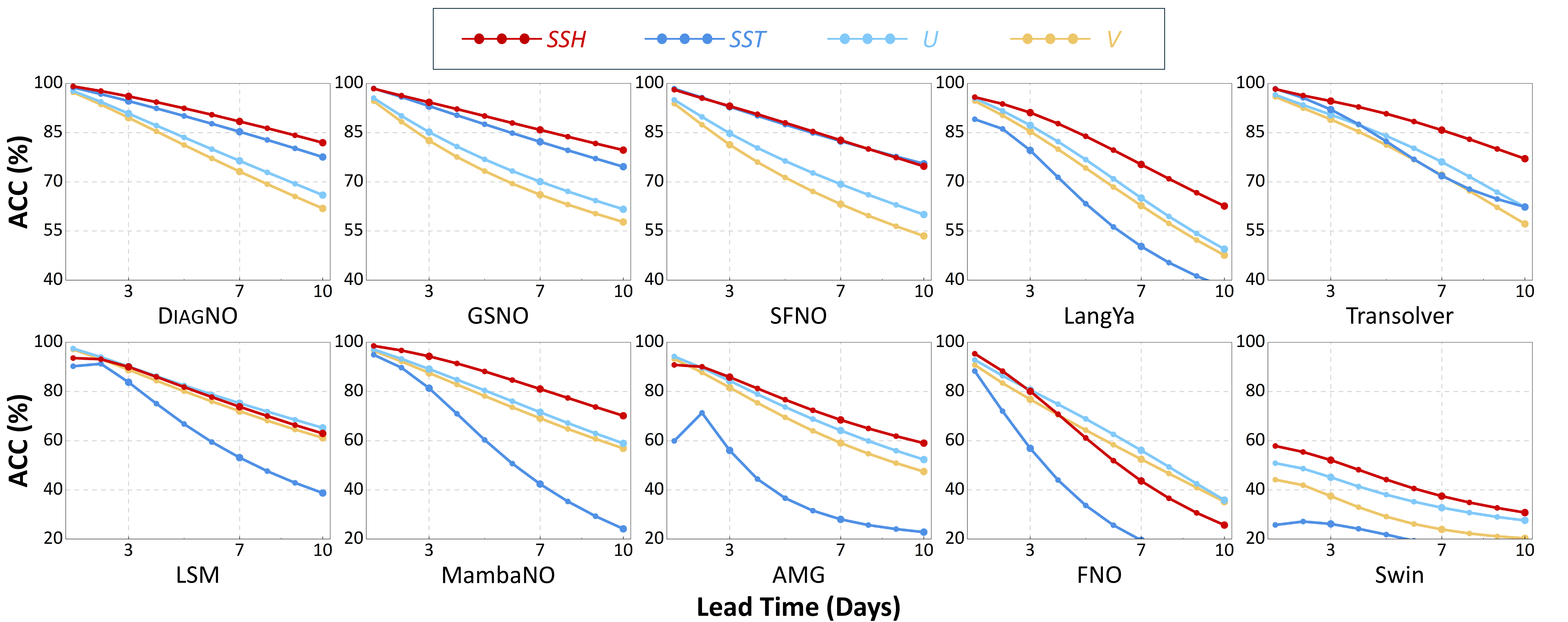}
        \caption{Daily Anomaly Correlation Coefficient (ACC) for four ocean surface variables ($SSH$, $SST$, $U$, $V$)}
        \label{app:glorys_page1}
    \end{subfigure}
    \vspace{1em}
    \begin{subfigure}{\textwidth}
        \centering
        \includegraphics[width=\linewidth, page=2]{imgs/appendix/Glorys_all.pdf}
        \caption{Daily prediction errors evaluated by MAE and RMSE.}
        \label{app:glorys_page2}
    \end{subfigure}
    
    \caption{Day-to-day forecasting performance of different models on global ocean surface variables over a 10-day lead time.}
    \label{app:glorys_all}
\end{figure}

Figure \ref{app:glorys_all} provides a day-to-day evaluation of the autoregressive forecasting stability of various models over a 10-day lead time. As illustrated by the Anomaly Correlation Coefficient (ACC) evolution for $SSH$, $SST$, $U$, and $V$ (Figure \ref{app:glorys_page1}), \textsc{DiagNO} demonstrates remarkable resilience against the inevitable correlation degradation typical of chaotic Earth system predictions. It exhibits the most gradual decay rate among all evaluated architectures, maintaining exceptionally high correlation near the 10-day mark, particularly for thermodynamically-driven variables like $SSH$ and $SST$. While standard spherical operators (GSNO, SFNO) show competitive short-term results, their accuracy declines more steeply over time. Conversely, general-purpose baselines (FNO, AMG, Swin) suffer severe performance drops. The sustained skill of \textsc{DiagNO} stems from its capacity to accurately resolve global-scale physical distributions (e.g., latitudinal climate belts) while explicitly preventing the generation and amplification of radial polar artifacts during recursive forecasting steps. This long-term stability is further corroborated by the absolute prediction errors, measured by Mean Absolute Error (MAE) and Root Mean Square Error (RMSE) (Figure \ref{app:glorys_page2}). As lead time increases, the error accumulation for \textsc{DiagNO} grows at a significantly slower and more suppressed rate compared to the baselines. In stark contrast, models like LangYa, FNO, and AMG exhibit an aggressive explosion in error after the third to fifth day, indicating a rapid breakdown in physical consistency during extended autoregressive inference. The tightly bounded error curves of \textsc{DiagNO} confirm that its internal representation effectively mitigates the exponential growth of numerical errors, establishing it as a highly reliable framework for continuous ocean state estimation.

\begin{figure}[htbp]
    \centering
    
    \begin{subfigure}{\textwidth}
        \centering
        \includegraphics[width=\linewidth, page=1]{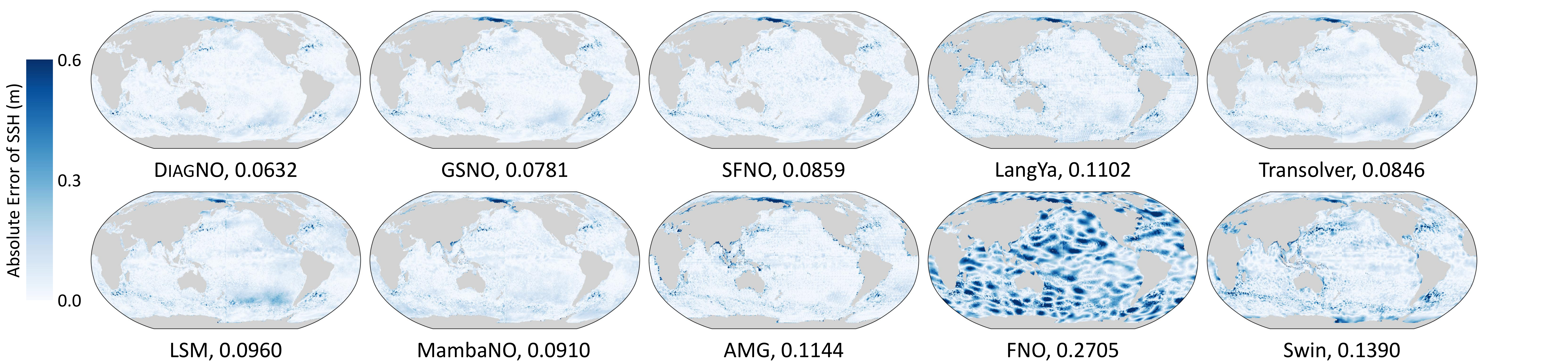}
        \caption{Absolute error visualization for Sea Surface Height ($SSH$).}
        \label{fig:glorys_vis_ssh}
    \end{subfigure}
    
    \vspace{1em}
    
    \begin{subfigure}{\textwidth}
        \centering
        \includegraphics[width=\linewidth, page=2]{imgs/appendix/Glorys_vis.pdf}
        \caption{Absolute error visualization for Sea Surface Temperature ($SST$).}
        \label{fig:glorys_vis_sst}
    \end{subfigure}
    
    \vspace{1em}
    
    \begin{subfigure}{\textwidth}
        \centering
        \includegraphics[width=\linewidth, page=3]{imgs/appendix/Glorys_vis.pdf}
        \caption{Absolute error visualization for Zonal Velocity ($U$).}
        \label{fig:glorys_vis_u}
    \end{subfigure}
    
    \vspace{1em}
    
    \begin{subfigure}{\textwidth}
        \centering
        \includegraphics[width=\linewidth, page=4]{imgs/appendix/Glorys_vis.pdf}
        \caption{Absolute error visualization for Meridional Velocity ($V$).}
        \label{fig:glorys_vis_v}
    \end{subfigure}
    
    \caption{Spatial visualizations of 10-day absolute prediction errors for four key ocean surface variables. The corresponding overall Root Mean Square Error (RMSE) for each evaluated model is explicitly annotated beneath its respective spatial error map.}
    \label{app:glorys_vis_all}
\end{figure}

Figure~\ref{app:glorys_vis_all} illustrates the spatial distribution of absolute prediction errors for all four ocean surface variables ($SSH$, $SST$, $U$, and $V$) at the 10-day lead time. The defining advantage of \textsc{DiagNO} lies in its robust capacity to significantly reduce error accumulation in universally challenging regions characterized by highly heterogeneous physical dynamics. For example, in the $SSH$ predictions, most standard baselines exhibit severe, concentrated error patches in the Arctic Ocean near Alaska. This specific region is dominated by complex sea ice-ocean interactions and sharp physical gradients along the marginal ice zone, which other models struggle to resolve accurately. \textsc{DiagNO} successfully mitigates these localized errors, maintaining physical consistency across such steep gradient zones. Furthermore, in the velocity channels ($U$ and $V$), massive error zones for baseline models are densely clustered along the highly dynamic equatorial current systems. \textsc{DiagNO} demonstrates notably lighter and cleaner error distributions in these complex circulatory regions. By effectively capturing multi-scale interactions and suppressing numerical instabilities in these notoriously volatile areas, \textsc{DiagNO} maintains superior global spatial coherence and achieves the most accurate long-term ocean state estimation.
\subsection{Atmospheric Forecasting}
\begin{figure}[htbp]
    \centering
    
    \begin{subfigure}{\textwidth}
        \centering
        \includegraphics[width=\linewidth, page=1]{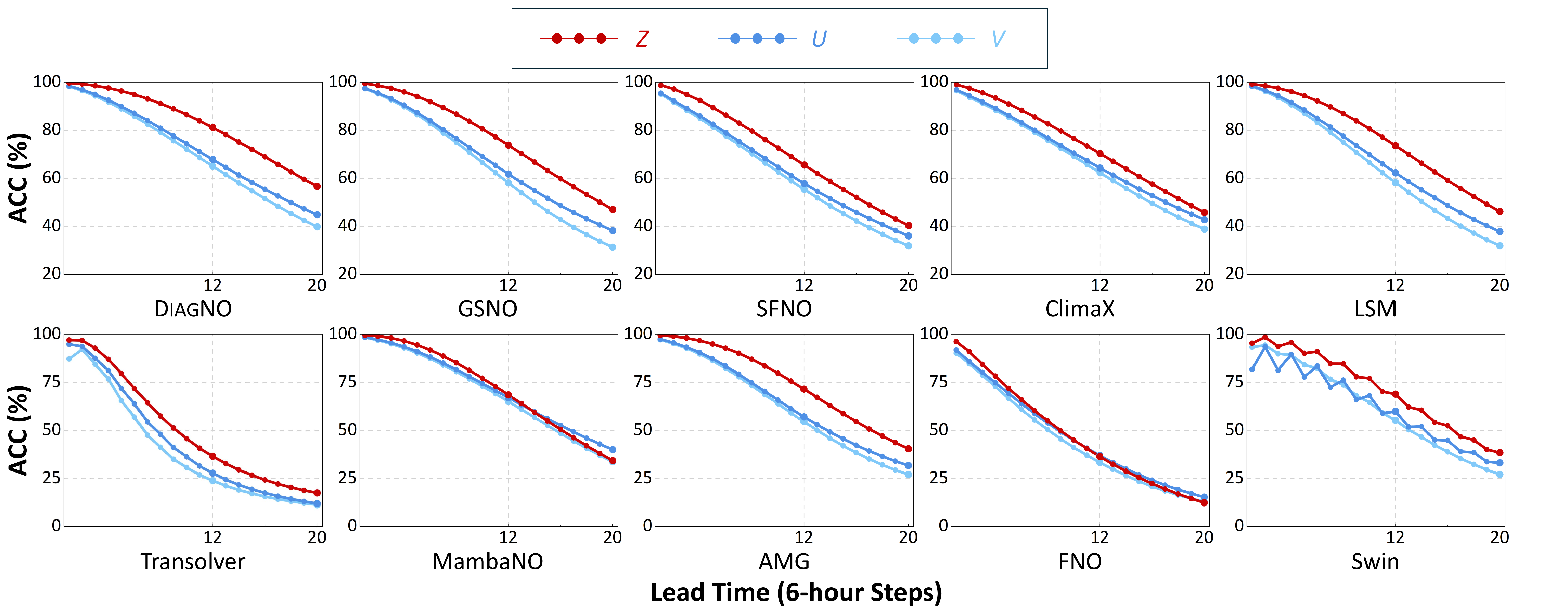}
        \caption{Step-wise Anomaly Correlation Coefficient (ACC) for three atmospheric variables ($Z$, $U$, $V$).}
        \label{app:era5_page1}
    \end{subfigure}
    
    \vspace{1em}
    
    \begin{subfigure}{\textwidth}
        \centering
        \includegraphics[width=\linewidth, page=2]{imgs/appendix/ERA5_all.pdf}
        \caption{Step-wise prediction errors evaluated by MAE and RMSE.}
        \label{app:era5_page2}
    \end{subfigure}
    
    \caption{Autoregressive forecasting performance of different models on global atmospheric variables over a 5-day lead time. The x-axis represents the forecast horizon spanning 20 consecutive 6-hour steps.}
    \label{app:era5_all}
\end{figure}
Figure \ref{app:era5_all} provides a detailed, step-wise evaluation of the autoregressive forecasting stability for three mid-troposphere (500 hPa) atmospheric variables ($Z$, $U$, and $V$) over a 5-day lead time. Since the model operates on a 6-hour temporal resolution, the x-axis represents 20 consecutive autoregressive steps.

Figure \ref{app:era5_page1} illustrates the degradation of the Anomaly Correlation Coefficient (ACC) as the forecast horizon extends. While predicting the chaotic dynamics of the atmosphere inherently leads to a decay in correlation skill, \textsc{DiagNO} exhibits the most gradual and resilient decay curve among all models. By step 20 (Day 5), \textsc{DiagNO} maintains the highest ACC scores across all three variables, particularly for the geopotential field ($Z$), which governs large-scale synoptic weather patterns. Advanced baselines like ClimaX, GSNO, and SFNO show competitive performance initially but degrade more rapidly in later steps. Notably, standard vision models like Swin display severe "zigzag" oscillations (sawtooth patterns) during the rollout, indicating severe temporal inconsistencies between consecutive 6-hour steps. \textsc{DiagNO}, however, produces exceptionally smooth and stable correlation trajectories, proving its structural superiority in maintaining global-scale atmospheric distributions over time.

Figure \ref{app:era5_page2} further validates this stability by tracking the accumulation of Mean Absolute Error (MAE) and Root Mean Square Error (RMSE). During the 20-step rollout, \textsc{DiagNO} consistently suppresses the error growth rate, culminating in the lowest absolute prediction errors at the 5-day mark. In contrast, models such as FNO, Transolver, and MambaNO experience a rapid, non-linear explosion in error after the initial few steps, reflecting a severe breakdown in physical consistency during multi-step inference. The tightly bounded error curves of \textsc{DiagNO} confirm its robustness in mitigating error accumulation, making it a highly reliable architecture for extended-range atmospheric forecasting.
\begin{figure}
    \centering
    
    \begin{subfigure}{\textwidth}
        \centering
        \includegraphics[width=\linewidth, page=1]{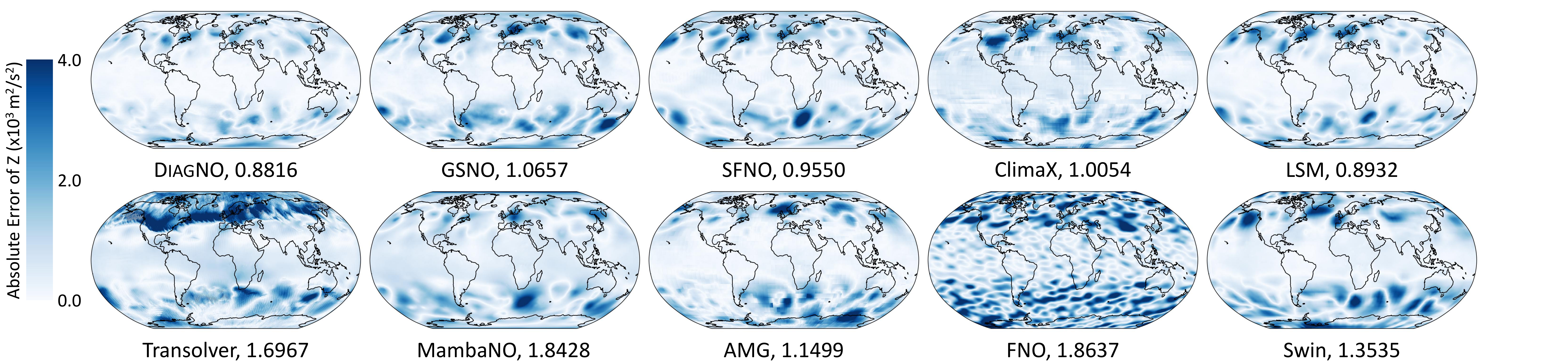}
        \caption{Absolute error visualization for Geopotential ($Z$).}
        \label{fig:era5_vis_z}
    \end{subfigure}
    
    \vspace{1em}
    
    \begin{subfigure}{\textwidth}
        \centering
        \includegraphics[width=\linewidth, page=2]{imgs/appendix/ERA5_vis.pdf}
        \caption{Absolute error visualization for Zonal Wind ($U$).}
        \label{fig:era5_vis_u}
    \end{subfigure}
    
    \vspace{1em}
    
    \begin{subfigure}{\textwidth}
        \centering
        \includegraphics[width=\linewidth, page=3]{imgs/appendix/ERA5_vis.pdf}
        \caption{Absolute error visualization for Meridional Wind ($V$).}
        \label{fig:era5_vis_v}
    \end{subfigure}
    
    \vspace{1em}
        
    \caption{Spatial visualizations of 20-step (5-day) absolute prediction errors for three key atmospheric variables. The corresponding overall Root Mean Square Error (RMSE) for each evaluated model is explicitly annotated beneath its respective spatial error map. Note that the annotated RMSE values for $Z$, $U$, and $V$ are scaled by a factor of $10^3$, $10$, and $10$, respectively, for visual clarity.}
    \label{app:era5_vis_all}
\end{figure}

Figure \ref{app:era5_vis_all} visualizes the spatial distribution of absolute prediction errors for Geopotential ($Z$), Zonal Wind ($U$), and Meridional Wind ($V$). Visually, \textsc{DiagNO} consistently exhibits the lightest error maps across all three variables, which perfectly aligns with its lowest overall RMSE. In contrast, standard models (e.g., FNO, Transolver) show significant error accumulations, particularly in high-latitude regions. While other spherical operators (e.g., SFNO, GSNO) improve upon this, they still exhibit noticeable regional errors in mid-latitudes. This direct comparison demonstrates that \textsc{DiagNO}'s diagonal interaction mechanism effectively captures global-scale heterogeneous dynamics, leading to superior spatial accuracy.
\subsection{Climate Anomaly}
\begin{table}[htbp]
  \centering
  \caption{RMSE evaluation of different models across ENSO phases.}
  \label{tab:enso_rmse}
  \begin{tabular}{lcccc}
    \toprule
    \multirow{2}{*}{MODEL} & \multirow{2}{*}{Total} & \multicolumn{3}{c}{RMSE} \\
    \cmidrule(lr){3-5}
    & & Neutral & El Ni\~no & La Ni\~na \\
    \midrule
    Swin       & 0.8713 & 0.2618 & 0.5053 & 0.1042 \\
    FNO        & 1.4281 & 0.4227 & 0.7389 & 0.2665 \\
    LSM        & 1.8451 & 0.4924 & 0.6296 & 0.7231 \\
    Transolver & 0.5427 & 0.1675 & 0.1376 & 0.2376 \\
    MambaNO    & 2.8335 & 0.8485 & 1.0422 & 0.9428 \\
    AMG        & 0.7727 & 0.2835 & 0.0953 & 0.3939 \\
    \midrule
    SFNO       & \underline{0.1771} & \underline{0.0871} & \textbf{0.0408} & \underline{0.0491} \\
    GSNO       & 0.3498 & 0.1058 & 0.1627 & 0.0813 \\
    \midrule
    LangYa     & 0.4008 & 0.1701 & 0.1517 & 0.0790 \\
    \midrule
    \textsc{DiagNO} & \textbf{0.1254} & \textbf{0.0579} & \underline{0.0535} & \textbf{0.0140} \\
    \bottomrule
  \end{tabular}
\end{table}
Table \ref{tab:enso_rmse} presents the detailed evaluation results for Sea Surface Temperature Anomalies (SSTA) at a 7-day lead time.
\newpage
\section{Impact Statement}
\label{impact statement}
This paper presents work aimed at advancing machine learning for Earth system modeling, with a specific focus on oceanic and atmospheric sciences. By providing efficient and high-fidelity forecasting for complex, global-scale ocean dynamics, we hope our research draws greater attention to these critical geophysical domains and accelerates vital downstream applications, such as marine monitoring and climate analysis. Ultimately, we aim to inspire further innovations that foster the continued flourishing of the broader AI for Science (AI4S) community. While our framework offers positive societal potential, we acknowledge the inherent limitations of purely data-driven methods—specifically, the risk of prediction failure when encountering unprecedented and extreme climate outliers. We therefore encourage the community to thoughtfully deploy such models as powerful augmentative tools alongside traditional physics-constrained systems.
\clearpage
\section*{NeurIPS Paper Checklist}
\begin{enumerate}

\item {\bf Claims}
    \item[] Question: Do the main claims made in the abstract and introduction accurately reflect the paper's contributions and scope?
    \item[] Answer: \answerYes{} 
    \item[] Justification: We discuss the contributions and scope in abstract and introduction.

\item {\bf Limitations}
    \item[] Question: Does the paper discuss the limitations of the work performed by the authors?
    \item[] Answer: \answerYes{} 
    \item[] Justification: We discuss the limitations in Section \ref{contributions & limitations}.

\item {\bf Theory assumptions and proofs}
    \item[] Question: For each theoretical result, does the paper provide the full set of assumptions and a complete (and correct) proof?
    \item[] Answer: \answerYes{} 
    \item[] Justification: Theoretical derivations are detailed in Section \ref{sec:method}, supported by rigorous proofs for modal isolation under rotational equivariance (Appendix \ref{proof:isolated_modes}) and optimal reconstruction via conjugate transpose (Appendix \ref{proof:best_approximation}).

    \item {\bf Experimental result reproducibility}
    \item[] Question: Does the paper fully disclose all the information needed to reproduce the main experimental results of the paper to the extent that it affects the main claims and/or conclusions of the paper (regardless of whether the code and data are provided or not)?
    \item[] Answer: \answerYes{} 
    \item[] Justification: We provide the training and inference strategies in Appendix \ref{training and inference}, dataset details in Appendix \ref{app:benchmarks}, and hyperparameter configurations in Appendix \ref{baselines}. Furthermore, the source code is publicly available on GitHub to support reproducibility.

\item {\bf Open access to data and code}
    \item[] Question: Does the paper provide open access to the data and code, with sufficient instructions to faithfully reproduce the main experimental results, as described in supplemental material?
    \item[] Answer: \answerYes{} 
    \item[] Justification: The source code is publicly available on GitHub, together with instructions for generating the simulated shallow water dataset. Due to copyright restrictions and massive file sizes, the atmospheric and oceanic reanalysis datasets are excluded. However, comprehensive details on dataset processing and variable discretization are explicitly provided in Appendix \ref{app:benchmarks}.

\item {\bf Experimental setting/details}
    \item[] Question: Does the paper specify all the training and test details (e.g., data splits, hyperparameters, how they were chosen, type of optimizer) necessary to understand the results?
    \item[] Answer: \answerYes{} 
    \item[] Justification: We detail the dataset splits in Section \ref{experiment setup} and discuss hyperparameter selection in Section \ref{ablation and hyperparameter}. Furthermore, the training methods, inference strategies, and optimizer configurations are explicitly described in Appendix \ref{training and inference}.

\item {\bf Experiment statistical significance}
    \item[] Question: Does the paper report error bars suitably and correctly defined or other appropriate information about the statistical significance of the experiments?
    \item[] Answer: \answerYes{} 
    \item[] Justification: As detailed in Section \ref{experiment setup}, we ensure statistical significance by conducting multiple independent runs for all experiments.

\item {\bf Experiments compute resources}
    \item[] Question: For each experiment, does the paper provide sufficient information on the computer resources (type of compute workers, memory, time of execution) needed to reproduce the experiments?
    \item[] Answer: \answerYes{} 
    \item[] Justification: We explicitly detail the computational resources required for our experiments in Appendix \ref{training and inference}.
    
\item {\bf Code of ethics}
    \item[] Question: Does the research conducted in the paper conform, in every respect, with the NeurIPS Code of Ethics \url{https://neurips.cc/public/EthicsGuidelines}?
    \item[] Answer: \answerYes{} 
    \item[] Justification: This research conforms the NeurIPS Code of Ethics.

\item {\bf Broader impacts}
    \item[] Question: Does the paper discuss both potential positive societal impacts and negative societal impacts of the work performed?
    \item[] Answer: \answerYes{} 
    \item[] Justification: We discuss the broader impacts in Appendix \ref{impact statement}, including both positive impacts and negative impacts.
    
\item {\bf Safeguards}
    \item[] Question: Does the paper describe safeguards that have been put in place for responsible release of data or models that have a high risk for misuse (e.g., pre-trained language models, image generators, or scraped datasets)?
    \item[] Answer: \answerNA{} 
    \item[] Justification: The paper poses no such risks.

\item {\bf Licenses for existing assets}
    \item[] Question: Are the creators or original owners of assets (e.g., code, data, models), used in the paper, properly credited and are the license and terms of use explicitly mentioned and properly respected?
    \item[] Answer: \answerYes{} 
    \item[] Justification: We properly cite all existing assets used in our research, including the ERA5 and GLORYS12 reanalysis datasets, as well as the open-source codebases for the baseline models. We strictly respect and adhere to their respective licenses and terms of use for academic research.

\item {\bf New assets}
    \item[] Question: Are new assets introduced in the paper well documented and is the documentation provided alongside the assets?
    \item[] Answer: \answerYes{} 
    \item[] Justification: We provide a README file in our public GitHub repository, offering instructions on environment configuration, hyperparameter adjustments, and experimental settings.

\item {\bf Crowdsourcing and research with human subjects}
    \item[] Question: For crowdsourcing experiments and research with human subjects, does the paper include the full text of instructions given to participants and screenshots, if applicable, as well as details about compensation (if any)? 
    \item[] Answer: \answerNA{} 
    \item[] Justification: This paper does not involve crowdsourcing or any research with human subjects.

\item {\bf Institutional review board (IRB) approvals or equivalent for research with human subjects}
    \item[] Question: Does the paper describe potential risks incurred by study participants, whether such risks were disclosed to the subjects, and whether Institutional Review Board (IRB) approvals (or an equivalent approval/review based on the requirements of your country or institution) were obtained?
    \item[] Answer: \answerNA{} 
    \item[] Justification: This research does not involve human subjects or any related human experiments.

\item {\bf Declaration of LLM usage}
    \item[] Question: Does the paper describe the usage of LLMs if it is an important, original, or non-standard component of the core methods in this research? Note that if the LLM is used only for writing, editing, or formatting purposes and does \emph{not} impact the core methodology, scientific rigor, or originality of the research, declaration is not required.
    \item[] Answer: \answerNA{} 
    \item[] Justification: The core method development in this research does not involve LLMs as any important, original, or non-standard components.

\end{enumerate}

\end{document}